\documentclass[10pt,twocolumn,letterpaper]{article}

\usepackage[pagenumbers]{cvpr} 
\usepackage{float}
\usepackage{xcolor} 
\usepackage[pagebackref,breaklinks,colorlinks,allcolors=cvprblue]{hyperref}
\usepackage{algorithm}
\usepackage{algpseudocode}
\usepackage{svg}
\usepackage{mdframed}
\usepackage{listings}
\usepackage{fancyvrb}
\newcommand{\selectionmethodname}{Batched Visual Selection}
\newcommand{\methodname}{FirePlace}
\newcommand{\supp}{Supp.~Mat.}

\newcommand{\red}[1]{{\color{red}#1}}
\newcommand{\green}[1]{{\color{teal}#1}}

\definecolor{cvprblue}{rgb}{0.21,0.49,0.74}

\definecolor{light-gray}{gray}{0.80}

\def\paperID{5756} 
\def\confName{CVPR}
\def\confYear{2025}

\title{
 \methodname: 
Geometric Re\underline{fi}nements of LLM Common Sense \underline{Re}asoning \\
for 3D Object \underline{Place}ment}

\author{
\textbf{Ian Huang}$^{1,2}$\footnotemark[1] \quad \textbf{Yanan Bao}$^{2}$ \quad \textbf{Karen Truong}$^{2}$ \\  
\textbf{Howard Zhou}$^{2}$ \quad \textbf{Cordelia Schmid}$^{2}$ \quad \textbf{Leonidas Guibas}$^{1,2}$ \quad
\textbf{Alireza Fathi}$^{2}$ \\[1em]
$^1$Stanford University \quad $^2$Google DeepMind \\[1em]
}

\begin{document}

\twocolumn[{%
\renewcommand\twocolumn[1][]{#1}%
\maketitle
\vspace{-3em}
\begin{center}
\centering
\captionsetup{type=figure}
    \includegraphics[width=0.95\textwidth]{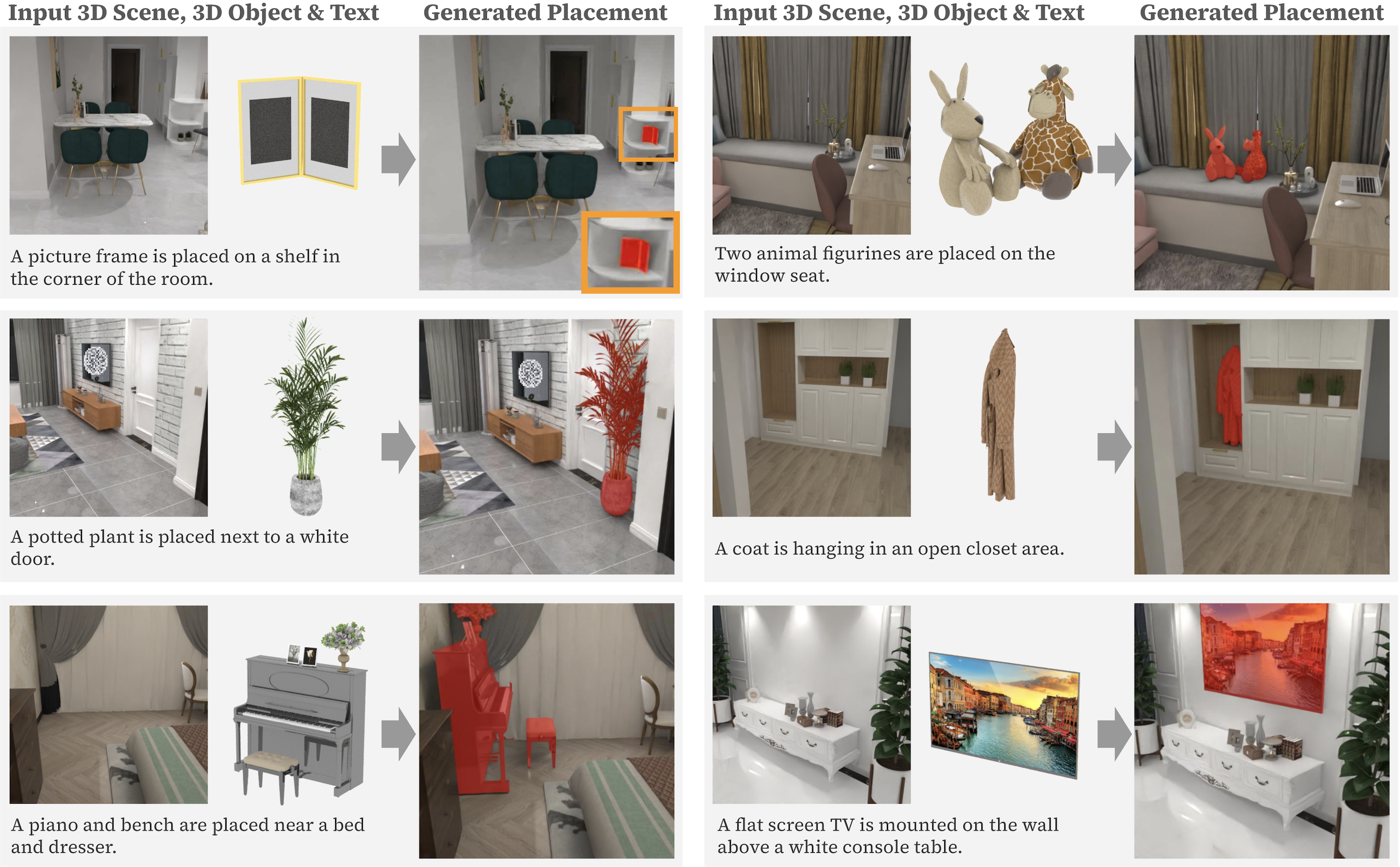}
    \caption{\methodname\ enables multi-modal large language models (MLLMs)to place new 3D objects into complex, preexisting 3D scenes, given (1) a 3D scene, (2) a 3D object, and (3) a language prompt. It uses a combination of MLLM common sense and low-level geometry constraints through the process described in this paper. Object placements generated by \methodname\ are shown in \textcolor{red}{Red}.}
    \label{fig:teaser}

\end{center}
}]

\maketitle
\footnotetext[1]{Work done during an internship at DeepMind. Correspondence to: \texttt{ianhuang@cs.stanford.edu}.}
\begin{abstract}
Scene generation with 3D assets presents a complex challenge, requiring both high-level semantic understanding and low-level geometric reasoning. While Multimodal Large Language Models (MLLMs) excel at semantic tasks, their application to 3D scene generation is hindered by their limited grounding on 3D geometry. In this paper, we investigate how to best work with MLLMs in an object placement task. Towards this goal, we introduce a novel framework, \methodname, that applies existing MLLMs in (1) 3D geometric reasoning and the extraction of relevant geometric details from the 3D scene, (2) constructing and solving geometric constraints on the extracted low-level geometry, and (3) pruning for final placements that conform to common sense. By combining geometric reasoning with real-world understanding of MLLMs, our method can propose object placements that satisfy both geometric constraints as well as high-level semantic common-sense considerations. Our experiments show that these capabilities allow our method to place objects more effectively in complex scenes with intricate geometry, surpassing the quality of prior work.

\end{abstract}

\section{Introduction}

Generating scenes with preexisting 3D assets is crucial for diverse applications like architecture, game development, and virtual reality. While constructing a scene can be seen as an iterative process of object placement, doing so even for a single object requires complex understanding of both high-level considerations for why an object should be placed at a particular location, as well as low-level understanding of the geometry of that object and its environment. 
Constructing such scenes requires placing objects in a way that is both physically feasible and contextually appropriate. This involves understanding high-level concepts like object relationships and aesthetics, as well as low-level geometric constraints.

Multi-modal Large Language Models (MLLMs) \cite{team2023gemini, 2023GPT4VisionSC} offer a promising approach due to their real-world knowledge and common-sense reasoning capabilities \cite{zhao2024large, li2021systematic}, which have been leveraged for tasks such as graphical editing \cite{huang2024blenderalchemy, 
yang2024holodeck, kulits2024re, wen2023anyhome, huang2023aladdin, feng2024layoutgpt, goel2023iterative}, scene understanding \cite{chandhok2024scenegpt}, and question answering \cite{yang2023dawnofllms, lee2024visual}. 
However, applying them to object placement leads to many failure cases, for the reason that MLLMs struggle with precise 3D spatial reasoning and fine-grained geometry \cite{el2024probing, you2024img2cad}. Existing efforts often resort to training or fine-tuning these models on extensive 3D datasets to enhance their spatial capabilities \cite{chen2024spatialvlm, li20243dmit, hong20233d,zhu2024empowering, huang2024reason3d, li2023m3dbench}.

This paper introduces \methodname, a novel approach for placing new 3D objects into existing scenes using \textit{off-the-shelf} MLLMs, guided by natural language instructions.
More concretely, our method takes as input a Universal Scene Descriptor (USD) of a 3D scene (including 3D assets within the scene), a camera angle capturing the scene, a new object mesh to be placed, and a language instruction describing the desired placement. \methodname\ then outputs an edited USD file, representing the scene with the object inserted. 

Our system iteratively translates abstract constraint descriptions (\eg, ``a book should be on the shelf'') into lower-level grounded 3D constraints (\ie \textit{Which part} of the book should be related to \textit{which part} of the shelf, and in \textit{what} way?). 
To accurately determine the parts of the scene that are relevant to the placement task, \methodname\ introduces a way for the MLLM to interrogate, visualize, and reason about relevant 3D information from the scene, by giving it access to a range of 3D processing tools. 
\methodname\ is able to combine the 3D processing capabilities of such tools with the common-sense reasoning capabilities of MLLMs to produce placements of objects that are \textit{feasible} (\ie, satisfy geometric constraints) and \textit{plausible} (\ie, satisfy common-sense reasoning about aesthetics, function and accessibility of object placements). \Cref{fig:teaser} shows some object placements generated by \methodname.

While recent works have introduced systems that generate object placements in scenes from scratch \cite{yang2024holodeck, hu2024scenecraft, aguina2024open, paschalidou2021atiss, wen2023anyhome, yang2024llplace, feng2024layoutgpt, tam2024scenemotifcoder, wei2023lego}, we note that they would perform poorly on the task of object placement in \textit{complex preexisting scenes} since they lack the following three capabilities:

\noindent\textbf{Reasoning with Fine-grained 3D Geometry.} Prior works in scene-generation propose methods that act upon bounding boxes of objects \cite{yang2024holodeck, huang2024blenderalchemy, hu2024scenecraft, aguina2024open, paschalidou2021atiss, yang2024llplace}, attending to coarse physical relationships between objects to be synthesized. However, such representations fail when the key surfaces for the placement task lie within the bounding box, as with shelves in \Cref{fig:teaser}. 
For example, placing a book on a shelf requires reasoning about the specific geometric features of the shelf's surface, such as its top plane and its depth, rather than simply its overall bounding box. 
To overcome the limitations of bounding-box-based approaches, \methodname\ reasons with fine-grained 3D geometry from an explicit 3D scene representation by extracting, visualizing, and reasoning about object surfaces. This allows it to generate scenes that are more realistic and physically plausible.

\noindent \textbf{Reasoning about object instances.} 
Existing scene generation methods  lack the ability to reference specific object instances (\eg \emph{which} chair, \emph{which} wall). These methods rely on pre-defined assumptions about wall layouts \cite{yang2024holodeck,aguina2024open,yang2024llplace, feng2024layoutgpt} and assume either that the specific instance choice is irrelevant or predetermined by construction~\cite{aguina2024open, yang2024holodeck, hu2024scenecraft}. This assumption fails when inserting objects into complex scenes, where the correct instance choice may be context-dependent. For example, hanging 
a picture in a room with multiple walls requires understanding the specific context to identify the intended wall. To tackle this, \methodname\ enables MLLMs to precisely reference object instances through \textit{visual} selection. 

\noindent \textbf{Common-sense placement.} 
Existing methods \cite{yang2024holodeck, hu2024scenecraft, feng2024layoutgpt, yang2024llplace, paschalidou2021atiss, wei2023lego} often neglect the final visual result and common-sense considerations such as aesthetics, functionality, and accessibility during object placement. 
\methodname\ addresses this limitation by leveraging the knowledge embedded in MLLMs, selecting among potential placements based on a wider range of criteria, including aesthetic appeal, functional appropriateness, and accessibility considerations. This results in more realistic and plausible scene configurations that go beyond mere geometric feasibility.

In order to enable MLLMs to reason about \textbf{fine-grained 3D geometry} and \textbf{object instances}, the MLLM must select among \textit{many} discrete choices (of objects, and of surfaces) to choose the right ones for the  placement task. However, MLLMs become substantially more prone to error as the set of options gets larger. We introduce a visual selection method that uses inference-compute scaling \cite{snell2024scalingllmtesttimecompute,brown2024largelanguagemonkeysscaling} to mitigate this, called \selectionmethodname, whereby choosing a single surface/object among a set of options is broken down into a multi-round decision process, and each round limits the number of options that the MLLM has to ``look at''. 
To summarize, our contributions are:
\vspace{3 pt}
\begin{enumerate}
    \item A 3D reasoning framework that enables an off-the-shelf MLLM to translate high-level understanding of object placement requests into geometrically grounded 3D constraint functions while also adhering to common sense. 
    \item \emph{\selectionmethodname:} 
    A method that enhances the reliability of MLLMs in complex visual selection tasks by increasing computational resources during inference.
    \item \emph{Experiments:} Our results show that \methodname\ surpasses existing LLM-based methods in generating realistic and plausible object placements within complex 3D scenes. Human evaluations confirm that \methodname\ produces placements that are both physically feasible and aligned with common-sense expectations.
\end{enumerate}
\vspace{3 pt}

\noindent Our investigation demonstrates the nuanced design choices and trade-offs needed when using MLLMs for downstream 3D tasks that require 3D understanding capabilities to be externally provided. 

\section{Related Works}
\noindent\textbf{3D scene generation and object placement.}
Significant efforts have been made towards 
collecting 3D scene datasets \cite{roberts2021hypersim, song2017suncg, song2015sunrgbd, chang2017matterport3d, fu20213dfront, raistrick2024infinigen}, enabling the community to train and develop systems that generate and/or position elements within indoor scenes \cite{wei2023lego, ma2018languagescenedatabases, paschalidou2021atiss,
ritchie2019fast, 
wang2019planit, 
wang2021sceneformer, fisher2012example, yang2024llplace}. While  
they demonstrate that object placement rules can be distilled from scene databases, these were not designed to handle open vocabularies of objects, and even less so to take into account the level of common sense reasoning that underlies human decisions to place objects where they are placed within our living environments. \methodname\ introduces a method to leverage the knowledge of MLLMs to do this in a training-free manner. While other works \cite{chen2023scenedreamer, henderson2024sampling, yi2024gaussiandreamer, chung2023luciddreamer, li2024art3d, yuan2024dreamscape, huang2024dreamwaltz, chen2024scalinggaussian, liu2021infinite, po2024compositional, ju2024diffindscene, lee2024semcity, hollein2023text2room, fridman2024scenescape, xie2024citydreamer, yu2024wonderjourney} use 2D image priors to generate scenes and objects, they often have issues preserving object identity  and physical plausibility of the final object arrangement. In contrast, \methodname\ works with an explicit 3D scene representation, where explicit geometric constraints are enforced.



\noindent\textbf{Foundation models for 3D graphics.}
More recent works~\cite{goel2023iterative, huang2024blenderalchemy, huang2023aladdin, wen2023anyhome, kulits2024re} have demonstrated the potential of involving large pretrained models for different stages of the 3D graphical design process. While they demonstrate capabilities in editing materials~\cite{huang2024blenderalchemy}, texture~\cite{huang2023aladdin}, and controlling animation \cite{goel2023iterative}, they struggle with tasks that require complex spatial reasoning, like object placement. Existing works like \cite{feng2024layoutgpt} have attempted to position objects in a scene by directly using LLMs through predicting the position and orientation of objects as LLM outputs. More recent works \cite{hu2024scenecraft, tam2024scenemotifcoder, aguina2024open, yang2024holodeck} have demonstrated the benefit of using LLMs to predict \textit{constraints} instead, before using a solver to solve for final object placements. However, despite being able to create 
large-scale scenes, they represent each object using bounding boxes, making it impossible to express fine-grained constraints between \textit{parts} of objects, leading to constraints that can only explain placements of box-like objects (as opposed to putting a book on a shelf, or a stuffed toy on a chair with a backrest and armrests). This design choice is understandable, since \textit{parts} of objects become increasingly hard for LLMs to reason about. \methodname\ introduces a way to overcome this limitation.


\section{Method}

\begin{figure*}[tb]
    \centering
    \includegraphics[width=0.9\textwidth]{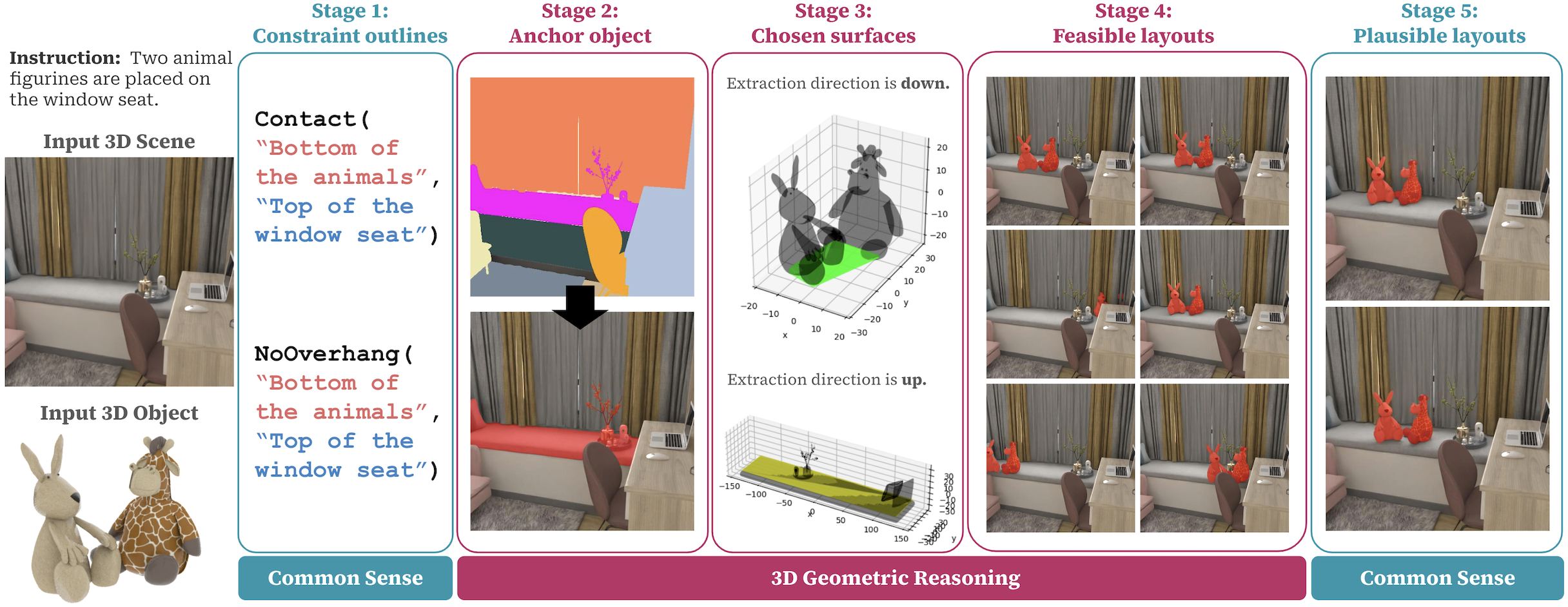}
    \caption{\methodname\ pipeline. \textbf{[Stage 1]} \methodname\ first generates a set of constraint outlines, describing in text from the applicable constraints and the corresponding interacting surfaces. \textbf{[Stages 2-4]} \methodname\ then selects the anchor object using \selectionmethodname\ on instance segmentation masks. It extracts the surfaces that best match the constraint outline, and then uses a constraint solver to produce feasible layouts. \textbf{[Stage 5]} Finally, it uses an MLLM to select a subset of placements that adhere to common sense principles.}
    \label{fig:problem_solving_process}
    \vspace{-1em}
\end{figure*}

As input, \methodname\ is given (1) a 3D scene $\mathcal D = \{o_1, o_2 ...\}$ defined as a collection of objects in world frame (along with a camera pose that captures the scene), (2)  a transformable 3D object $O_T$ to be inserted, and (3) a text instruction $l$ that describes the desired placement of the object. \methodname\ returns a single transformation matrix $T$, for the final placement of the object $T(O_T)$ in the world frame. We desire a $\mathcal D' = \mathcal D \cup \{T(O_T)\}$ that  best matches the description $l$.

\methodname\ can be seen as executing a visual chain-of-thought process that breaks down the visual reasoning task of object placement into individual steps. \methodname\ starts by leveraging MLLMs' world knowledge to generate ``constraint sketches'', which is a representation that references a list of constraint functions, and textually annotates the surfaces that these constraints should act upon. \methodname\ then performs three 3D reasoning stages of (1) resolving the surface references (\eg ``the seat of the chair'') to the 3D surfaces of the objects, (2) estimating continuous parameters, if applicable, of the constraint functions chosen in the constraint sketch, and (3) using a constraint solver to produce candidate object placements that satisfy the constraints. \methodname\ finally prunes the set of candidate object placements based on their renderings, selecting ones that adhere the most to common-sense reasoning of aesthetics, functionality and accessibility. \Cref{fig:problem_solving_process} shows the outcome of these stages in action.

\subsection{Constraint Outline Generation [Stage 1]} \label{sec:constraint_outline}

Given text $l$, constraint library $\mathcal F$ (described in \Cref{sec:constraint_lib}), and the rendering $H_R(\mathcal D)$ of the scene (for some rendering function $H_R$), the MLLM is prompted to produce a list $G$ of triplets $(f, t_A, t_T)$, where $f \in \mathcal F$ 
references the specific constraint that should be followed, and $t_A$ and $t_T$ 
are text descriptions describing the relevant surfaces \textit{participating in this constraint}. $t_A$ corresponds to the surface of the anchor object (\eg ``the \textit{top} surface of the white cabinet) and $t_T$ to that of the transformable object (\eg the bottom of the TV screen). Generating constraint outlines can be accomplished simply by appropriately prompting the MLLM and parsing its results, which we will show in the \supp. An example of this can be seen in Stage 1 of \Cref{fig:problem_solving_process}.

\subsection{3D Reasoning for Feasibility [Stages 2-4]} \label{sec:feasibility_reasoning}

In the 3D reasoning stage, the objective is to form placement candidates $\{T_{i}\}$  that satisfy geometric constraints expressed by the constraint outline. In order to do so, \methodname\ must resolve the references of $t_A$ and $t_T$ \textit{down to the level of 3D surfaces}, for every triplet within outline $G$. 

The method is described in \Cref{alg:constraint_construction}. For each triplet, the method starts by running a visual selection process among the objects within the scene $\mathcal D = \{o_1, o_2 ... \}$ for the  object, $O_A$, that best matches the description  $t_A$. \methodname\ renders the segmentation masks of every object visible in the camera viewpoint, $H_S(o_i) $, and tasks the MLLM with ``pointing'' to the object by naming the  color of the associated segmentation mask. This enables \methodname\ to \textbf{reason about object instances}. See Stage 2 of \Cref{fig:problem_solving_process}.

\begin{algorithm}
\caption{3D Reasoning of \methodname}
\label{alg:constraint_construction}
\begin{algorithmic}[1]
\Procedure{Construct Constraints}{
3D Scene $\mathcal D$,  Target Object $O_T$, Constraint outline $G$
}


\State $\text{Constraints } R \gets \emptyset$
\For {$(f, t_A, t_T) \text{ in } G$}
    \State $ O_A \gets V( \mathcal D, t_A)$ \green{\Comment{Choose anchor object}}
    \State $ \vec d_A \gets \text{DirExtr(} O_A, t_A \text{)}$ \green{\Comment{$O_A$ surface normal}}
    \State $ \vec d_T \gets \text{DirExtr(} O_T, t_T \text{)}$ \green{\Comment{$O_T$ surface normal}}
    \State $\{s^A_1, ...\} \gets \text{SurfExtr(}  
O_A, \vec d_A \text{)}$ \green{\Comment{$O_A$ surfaces}}
    \State $\{s^T_1, ...\} \gets \text{SurfExtr(}
O_T, \vec d_T \text{)}$ \green{\Comment{$O_T$ surfaces}}
    \State $s^A_* \gets V( \{s^A_1, ... \} , t_A)$ \green{\Comment{Choose $O_A$ surface}}
    \State $s^T_* \gets V( \{s^T_1, ...\} , t_T)$ \green{\Comment{Choose $O_T$ surface}}
    \State $\pi \gets \text{ContinuousParam(} f \textbf{)}$ \green{\Comment{Guess params}}
    \State $ R \gets R \cup \{(f, s^A_*, s^T_*, \pi)\}$ \green{\Comment{Add to constraints}}
\EndFor
\State $ \{T^*_1 ...\}\gets \arg\min_{T} \sum_{(f, s^A, s^T , \pi) \in R} f(s^A, T(s^T), \pi)$ \green{\Comment{Solve for feasible set of transformations}}
\State \Return $\{T^*_1, ...\}$
\EndProcedure
\end{algorithmic}
\end{algorithm}

Then, \methodname\ renders $H_R(O_A)$, and tasks the MLLM in $\text{DirExtr}(.)$ to extract the \textit{normal facing direction} for the surfaces of $O_A$ that best correspond with $t_A$. For instance, if $t_A$ mentions the \textit{seat} of the chair,  $\text{DirExtr}()$ will return $\verb|up|$. The same is done for $O_T$  and $t_T$. \methodname\ then uses geometric processing algorithms (described further in the \supp) to extract sets of surfaces from both $O_T$ and $O_A$ based on their face-normals and their alignment with the surface directions extracted previously. This gives us a set of candidate interaction surfaces $\{s^{A}_1...\}$ for $O_A$ and $\{s^{T}_1...\}$ for $O_T$. Each surface is expressed as a planar convex hull in 3D, circumscribing the faces that match. 

It's expected that within a complex 3D asset, there will be many candidate interaction surfaces for any particular direction. \methodname\ therefore uses a visual selection process once again to choose the best match among them ($s_*^T$ and $s_*^A$) corresponding to $t_T$ and $t_A$. \methodname\ renders the interaction surfaces $\{s^A_1...\}$ and $\{s^T_1...\}$, overlaid on top of $O_T$ and $O_A$, and renders these surfaces in different colors. It then tasks the MLLM to ``point'' to the best plane by naming its color. This allows \methodname\ to express constraints in terms of \textbf{fine-grained geometry.} See Stage 3 of \Cref{fig:problem_solving_process}.

Finally, the MLLM is prompted to estimate continuous arguments $\pi$ (\eg  such as distances for distance constraints), completing the last piece of a fully 3D-grounded constraint function (\supp\ shows the prompts used). This function can then be evaluated for some transformation $T$ as $f(s^A_* , T(s^T_*) , \pi)$. After  accumulating this for every outline in $G$, \methodname\ minimizes the sum of the constraint functions with respect to $T$ using simulated annealing,  as done in \cite{raistrick2024infinigen}, and returns the set of candidate placements that successfully minimize the sum of the constraint functions below a scalar threshold. See Stage 4 of \Cref{fig:problem_solving_process}.

\subsection{Plausibility Pruning [Stage 5]}

As done in \cite{huang2024blenderalchemy}, we use MLLM-based pruning of the final results to remove placements that are \textit{geometrically feasible} but \textit{implausible} based on considerations such as aesthetics, functionality, and accessibility. This is done  by running visual-selection on the renderings $H_R(D \cup \{T_i^*(O_T)\})$ through tasking the MLLM to find the most plausible placements. This stage guides the MLLM to first generate considerations for what \textit{good} placements would look like before picking between pairs of renderings shown. By the end of this stage, the selected placement adheres more to \textbf{common sense}. See Stage 5 of \Cref{fig:problem_solving_process}.

\subsection{Constraint Library}\label{sec:constraint_lib}
We provide \methodname\ with a small constraint library composed of binary constraint functions. They take in two interaction surfaces (from the anchor and target objects) and output a loss, which is minimal (\ie $0$) when the constraint is met.
We provide \methodname\ with the following constraint functions:
(1) $\verb|Parallel(p1, p2)|$: $\verb|p1|$ and $\verb|p2|$ are parallel.
(2) $\verb|CloseTo(p1, p2, dist)|$: the maximum distance between $\verb|p1|$ and $\verb|p2|$ is $\verb|dist|$.
(3) $\verb|FarFrom(p1, p2, dist)|$: the minimum distance between $\verb|p1|$ and $\verb|p2|$ is $\verb|dist|$.
(4) $\verb|InfrontPlane(p1, p2)|$: $\verb|p2|$ is on the ``front-facing'' side of $\verb|p1|$ (in the direction of its surface normal).
(5) $\verb|Contact(p1, p2)|$: $\verb|p1|$ and $\verb|p2|$ are in contact.
(6) $\verb|NoOverhang(p1, p2)|$: $\verb|p1|$'s projection onto $\verb|p2|$ is entirely contained within $\verb|p2|$. This is useful for expressing that an object placed on another should not overhang the edge of the top surface it sits upon.

While this library is small, we find that this set of planar  constraint functions can, in combinations, communicate most, if not all, of the constraint functions in prior works like \cite{yang2024holodeck, hu2024scenecraft}. We provide more details in the \supp. 

\subsection{\selectionmethodname} \label{sec:visual_selection}

\begin{algorithm}
\caption{\selectionmethodname}
\label{alg:visual_selection_algo}
\begin{algorithmic}[1]
\Procedure{V}{candidates $\{c_1, c_2, ...c_k\}$, objective $t$, batch size $m$}

\State $C \gets [c_1, c_2, ... c_k]$ \green{\Comment{List of candidates}}
\If{$|C| < m$}
    \State \Return $\text{Best match among } C \text{ according to } t$
\EndIf
\While{$|C| > 1$} \green{\Comment{If multiple candidates exist}}
    \State $C \gets \text{Shuffle}(C)$ \green{\Comment {Shuffle before batching}}
    \State $C \gets \bigcup_{i=0}^{|C|/m-1} V(\{C_{im+1} ... C_{m(i+1)}\}, t, m)$
    \green{\Comment{Break into m-sized batches then recursively call self.}}
\EndWhile
\State \Return $C$
\EndProcedure
\end{algorithmic}
\end{algorithm}

\methodname\ uses visual selection pervasively throughout its 3D reasoning and plausibility pruning. However, selecting among many visual options is a challenge for MLLMs. Similar to findings in \cite{wu2312v}, we found that overwhelming the MLLM with visual options heavily impacts its ability to choose the correct object or interaction surfaces.


Inspired by the use of inference compute scaling to overcome inherent limitations in MLLMs \cite{huang2024blenderalchemy, brown2024largelanguagemonkeysscaling, snell2024scalingllmtesttimecompute}, we propose \selectionmethodname\ (outlined in \Cref{alg:visual_selection_algo}), a method that recursively breaks down the set of options into batches of size $m$. The method shows at most $m$ options to the MLLM at a time, merges the ones that get selected within each batch, and repeats this process until a single option remains. 

In the context of visual-selection for anchor objects, $m$ candidate objects within a certain batch are masked out from the scene rendering in $m$ different colors. The MLLM is provided a list of these color names alongside the image, and selects options by referencing color names. Similarly, $m$ candidate interaction surfaces within a mask are visualized on top of the original object mesh in $m$ different colors, and the MLLM is prompted to choose the most relevant interaction surface in a similar fashion.

\section{Experiments}

\begin{figure*}[tb]
    \centering
    \includegraphics[width=0.96\textwidth]{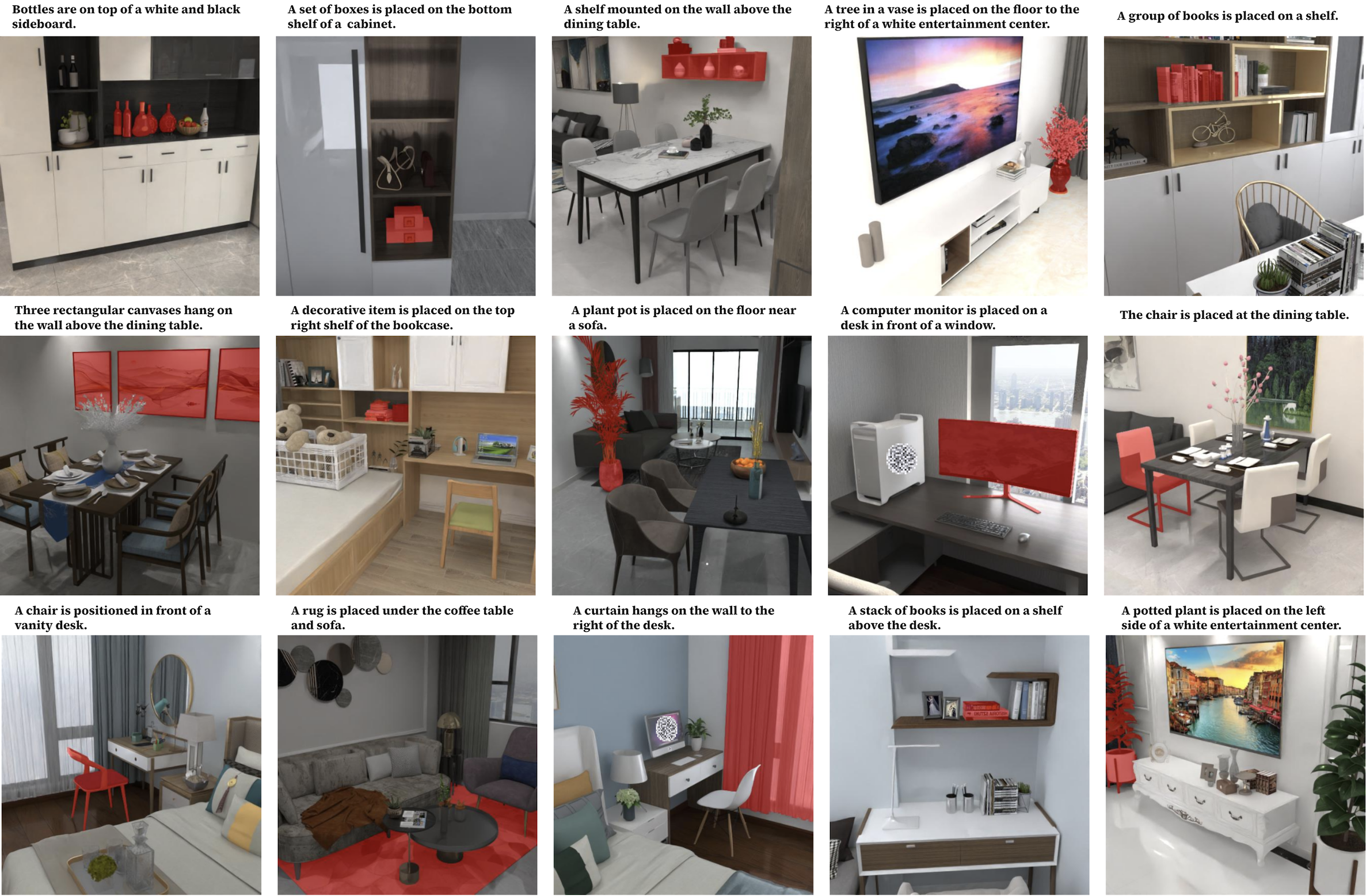}
    \caption{Qualitative samples of object placements (shown in \red{red} masks) within 3D scenes based on language instructions. \methodname\ can place diverse objects in a variety of settings, and produce geometrically feasible and semantically plausible object placements.}
    \label{fig:qualitative_samples1}
    \vspace{-1em}
\end{figure*}

\subsection{Dataset, Metrics \& Baselines}

To demonstrate \methodname's ability to perform object placement in complicated scenes, we  conduct our experiments on 50 photorealistic 3D scenes with fixed camera viewpoints designed by human experts. To create placement tasks, we select one object for each placement task (transformable object $O_T$) and caption its placement in the scene using language annotation $l$. We then remove the object from the scene and save the modified scene as the initial scene $\mathcal D$. Doing so for 
a number of objects in each of the 50 scenes creates an evaluation dataset of 266 placement tasks.
More details on these 3D scenes and tasks can be found in the \supp. 

We use five metrics to evaluate our placements:
\begin{enumerate}
\item \textbf {Min L2 Error}: L2 error of the closest match to the groundtruth translation among multiple tries per placement task, generated by each method.
\item \textbf {Mean L2 Error}: L2 error in the groundtruth translation and the predicted translation of the object, averaged across multiple tries per placement task.
\item \textbf{Energy Score}: the proportion of constraint functions generated by the constraint generator that outputs LOW energy at the groundtruth placements ( 
$<0.01$). 
\item \textbf{Plausibility Score}: Inspired by the latest works that use MLLMs for evaluation of 3D generative models \cite{wu2024gpt}, the plausibility score is generated by Gemini, and ranges from 1 (poor) to  4 (great). We provide it a concrete rubric  for evaluation of every  sample, which is shown in detail in the \supp. \Cref{sec:human_eval} also shows great alignment between the plausibility score and human preferences in our user preference study.
\item \textbf{Visibility Score} (in range [0,1]) is the rate which transformable objects are observable within the generated placement renderings after placement.
\end{enumerate}

We compare against Holodeck\cite{yang2024holodeck} and LayoutGPT\cite{feng2024layoutgpt}, works that use LLMs for the generation of constraints of object bounding boxes \cite{yang2024holodeck} or 3D position of object bounding boxes \cite{feng2024layoutgpt}. For both methods, we provide a list of Gemini-generated captions for the objects within the scene (generated based on per-object renderings) and ask their constraints/positions to be generated in the context of preexisting objects in the scene. Our full prompts for both methods are shown in the \supp. For Holodeck, we only use the ``Constraint-based Layout Design Module'' \cite{yang2024holodeck}, the module relevant to placement.

We choose to compare with both methods because both leverage the common-sense reasoning of large pretrained foundation models, but are lacking in the 3 capabilities enabled by \methodname\ in different ways. For our method and in all baselines, Gemini-1.5 Pro \cite{team2023gemini} is used. 

\subsection{Qualitative Examples \& Baseline Comparisons}

\begin{figure*}[tb]
    \centering
    \includegraphics[width=0.85\textwidth]{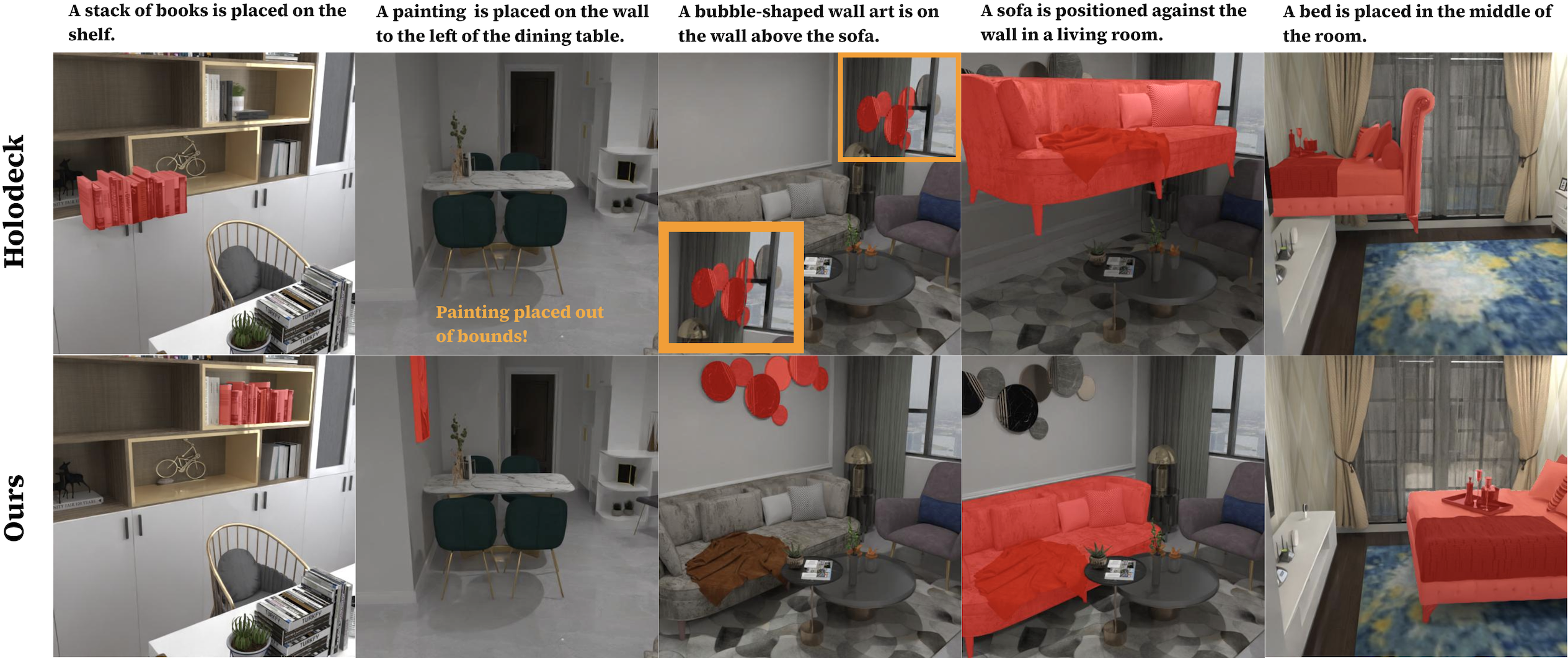}
    \caption{\textbf{Comparisons against Holodeck.} Holodeck fails to put the collection of books onto the shelf (due to its bounding box representation), and produces many implausible placements due to incorrect selection of anchor objects using the caption-based selection method.}
    \label{fig:qualitative_vsholodeck}
    \vspace{-1em}
\end{figure*}


\begin{table}[tb]
    
    \centering
    \begingroup
    \setlength{\tabcolsep}{5pt} 
    \renewcommand{\arraystretch}{1.} 
    \begin{tabular}{@{\hskip 0.5em}rcccc@{\hskip 0.5em}}
    \toprule
    Metric & LayoutGPT & Holodeck & Ours \\
    \midrule 
Min L2 error cm ($\downarrow$) & 89.01  & 113.17  & \textbf{48.39} \\
Mean L2 error cm ($\downarrow$) & 132.96 & 137.60  & \textbf{69.89}  \\
Visibility score ($\uparrow$) & 0.79   & 0.63    & \textbf{0.88}   \\
Plausibility score ($\uparrow$) & 2.14   & 2.13    & \textbf{2.95}   \\
Energy score ($\uparrow$) & N/A & 0.22    & \textbf{0.42}   \\
    \bottomrule
    \end{tabular}
    \caption{Comparison with Holodeck\cite{yang2024holodeck} and LayoutGPT\cite{feng2024layoutgpt}. LayoutGPT does not use constraints, so Energy Score is N/A. \methodname\ outperforms both baselines in all of the 5 metrics. See \Cref{fig:qualitative_vsholodeck} and \Cref{fig:qualitative_vslayoutgpt} for visual samples.}
    \label{tab:text_baseline_comparisons} 
    \endgroup
    \vspace{-1em}
\end{table}

Like \Cref{fig:teaser}, \Cref{fig:qualitative_samples1} shows the breadth of objects that can be placed into a variety of existing 3D scenes using \methodname. This demonstrates \methodname's capabilities in producing object placements compatible with visual context and lower-level geometry like the surfaces of bookshelves.

\Cref{fig:problem_solving_process} shows the intermittent steps of placing animal figurines on the window seat. Note that for the figurines to be put onto the window seat, the 3D reasoning stage must both select the window seat from the  many segmentation masks in the image \textit{and} extract the surface that corresponds to \textit{placeable} parts of the window seat. Despite the fact that cushions and plant vases are \textit{attached} to the window seat mesh, \methodname\ extracts and chooses the  placeable surface. Additionally, note that among the feasible placements, a few are aesthetic/accessible positions. Plausibility pruning is able to select more centered placements among these as the final output. 

\begin{figure*}[tb]
    \centering
    \includegraphics[width=0.85\textwidth]{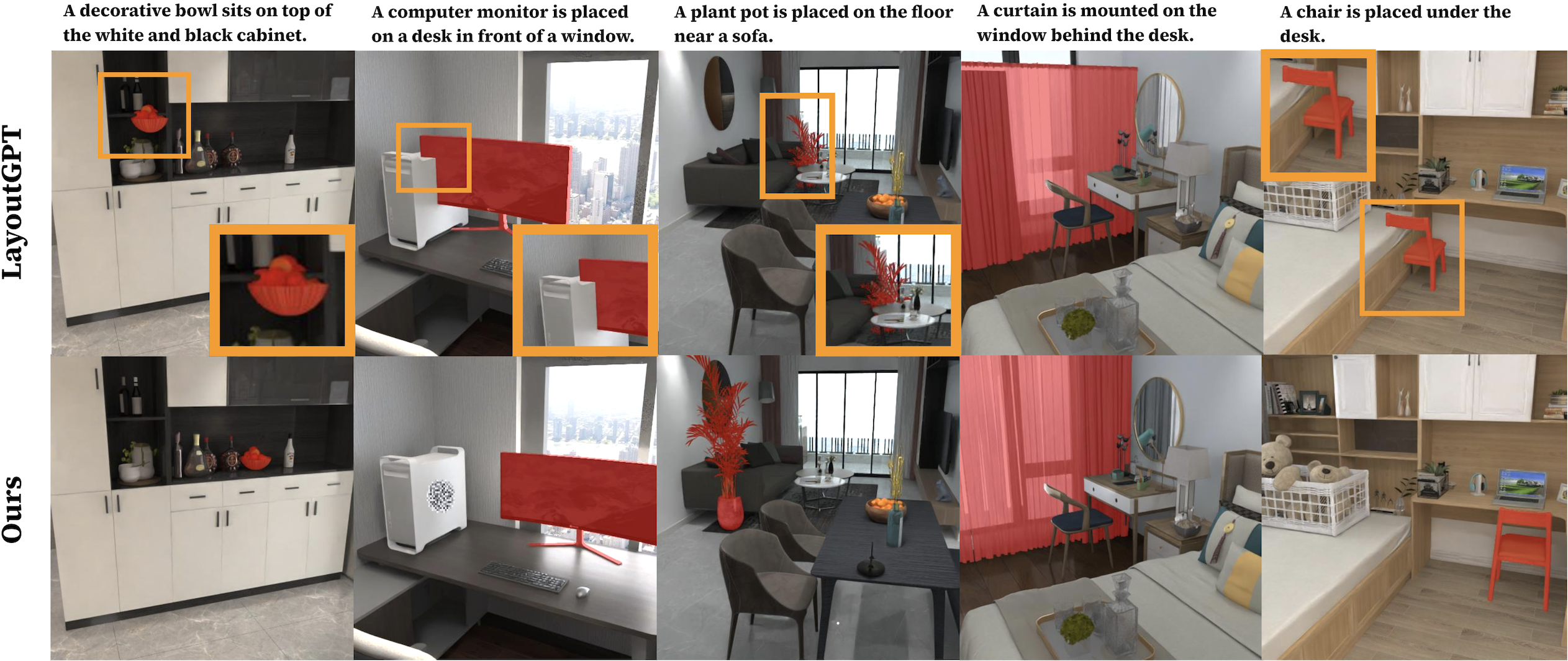}
    \caption{\textbf{Comparisons against LayoutGPT.} LayoutGPT produces implausible object placements with intersections, showing that LLMs often fail to accurately estimate object positions and should be guided by constraints, as done in \methodname.}
    \label{fig:qualitative_vslayoutgpt}
    \vspace{-1.5em}
\end{figure*}

\Cref{tab:text_baseline_comparisons} shows a comparison of our method against Holodeck and LayoutGPT across the 266 placement tasks. 
All methods are given language descriptions of the desired placement. \methodname\ achieves close to \textit{half} the L2 losses of baselines, and additionally scores higher on visibility, plausibility, and energy. Qualitatively comparing outputs (\Cref{fig:qualitative_vsholodeck} and \Cref{fig:qualitative_vslayoutgpt}) shows that our  method is able to produce higher quality placements.


\subsection{Human Evaluations} \label{sec:human_eval}
\begin{table}[tb]
    \centering
    \begingroup
    \setlength{\tabcolsep}{5pt} 
    \renewcommand{\arraystretch}{1.} 
    \begin{tabular}{@{\hskip 0.5em}cccc@{\hskip 0.5em}}
    \toprule

    & Physics & Semantics & Com. Sense \\
    \midrule 
    Win vs. Holodeck & \textbf{60.27\%} & \textbf{65.70\%} & \textbf{62.08\%} \\
    Tie vs. Holodeck & 19.69\% & 20.77\% & 19.81\% \\
    \midrule 
    Win vs. LayoutGPT & \textbf{76.02\%} & \textbf{76.10\%} & \textbf{76.42\%} \\
    Tie vs. LayoutGPT & 11.39\% & 11.47\% & 11.22\% \\

    \bottomrule
    \end{tabular}
    \caption{User preference study shows that \methodname\ placements are preferred over both baselines along all 3 axes of comparisons.}
    \label{tab:human_evaluation}
    \endgroup
    \vspace{-1em}
\end{table}

We perform user preference studies (30 participants, 
2075 side-by-side comparisons
to LayoutGPT and Holodeck) on randomly chosen outputs of \methodname, LayoutGPT, and Holodeck.
Human participants are shown masked renderings of two placements (presented as option A and B in random order) and the original language prompt.
Participants are tasked with annotating which of two placements (or tie) is (1) more physically plausible (\eg objects are not floating in space), (2) more semantically aligned with the input language prompt, and (3) more plausible according to common sense -- with considerations for aesthetics, functionality, and accessibility within living spaces. We show the results in \Cref{tab:human_evaluation} for the win-rate of outputs from our method compared to baselines for questions (1), (2) and (3) under ``physics'', ``semantics'' and ``com. sense'' respectively. We surpass the performance of prior works quite notably, and predominantly tie in human preferences when our method does not clearly produce a better result, as judged by our human annotators. These human annotations also show a high correlation with \textit{plausibility}. Among the cases when a clear winner is decided by human raters, plausibility metrics and human judgment agree \textbf{89.82\%} of the time. 

\subsection{Ablation Studies}
\begin{table}[t]
    \centering
    \begingroup
    \setlength{\tabcolsep}{4pt} 
    \renewcommand{\arraystretch}{1.2} 
    \begin{tabular}{@{\hskip 0.5em}rcccc@{\hskip 0.5em}}
    \toprule
    Ablation & L2 ($\downarrow$) & Vis($\uparrow$) & Plaus.($\uparrow$) & Energy ($\uparrow$) \\
    \midrule
    Ours                  & \textbf{62.52}  & \textbf{0.87}  & \textbf{2.94}  & 0.34  \\
    $-$ Constraints       & 124.11 & 0.24  & 1.95  & N/A     \\
    $-$ Vis. Select.      & 67.61  & 0.82  & 2.85  & 0.25  \\
    $-$ Geometry   & 63.44  & 0.79  & 2.89  & \textbf{0.44}  \\
    $-$ Com. Sense       & 66.08  & \textbf{0.87}  & 2.60  & 0.34  \\
    $-$ Vis. Scale.       & 83.95  & 0.84  & 2.55  & 0.19  \\
    \bottomrule
    \end{tabular}
    \caption{Ablations of design decisions in \methodname. L2, Vis, Plaus, and Energy are short for mean L2 error, visibility, plausibility, and energy score, respectively.}
    \label{tab:IMAGE_Ablations}
    \endgroup
    \vspace{-1em}
\end{table}



\methodname\ introduces many design choices that we validate through ablations. The results are shown in \Cref{tab:IMAGE_Ablations}.

\noindent \textbf{Use of geometric constraints.} To what extent can the placement problem be solved by asking the MLLM to pick the best placement among random guesses of placement? The ``$-$ Constraints'' row in \Cref{tab:IMAGE_Ablations} demonstrates the consequences of doing this -- \methodname\ generates 100 random placements, renders them all, and asks the VLM to choose among them. The result has substantially worse L2 error, visibility (notably, placed objects are only visible 24\% of the time), and plausibility scores, motivating the need for a method that uses fine-grained geometric constraints to guide the common-sense selection. 

\noindent \textbf{Visual selection of anchor objects.} What happens when we remove the visual selection process from \methodname, and use caption-based selection (as done for the baselines) instead? The results are shown in the ``$-$ Vis.~Select.'' row in \Cref{tab:IMAGE_Ablations}. Removing the ability to visually select object instances for the anchor objects heavily relies on  robust captioning for the assets, leading to drops in performance across the board. Moreover, the generated constraints match groundtruth placements less well, as shown by the substantial decrease in energy score.

\noindent \textbf{Fine-grained geometry.} 
We replace interaction surfaces extracted in \Cref{alg:constraint_construction} 
with the bounding box surfaces facing that direction in row ``$-$Geometry'' of \Cref{tab:IMAGE_Ablations}.
Interestingly, we find that the energy score becomes higher, likely due to the fact that among the fewer constraints  predicted by \methodname, many \textit{are} explainable using bounding box constraints. However, this leads to a lower expressivity of the constraints, evidenced by the dip in plausibility and a higher chance of the object being placed out of view, shown by the lower visibility score.
Qualitatively, this leads to the observation that objects are often unable to be put onto shelves or flat surfaces that are circumscribed by bounding boxes substantially larger than it or at a different elevation.

\noindent \textbf{Plausibility pruning.} Row ``$-$ Com. Sense'' in \Cref{tab:IMAGE_Ablations} shows the consequences of simply using the geometrically feasible placement solutions as final solutions, without plausibility pruning. This causes a large dip in plausibility, leaving visibility and energy scores largely unchanged.

\noindent \textbf{Inference compute scaling for \selectionmethodname.} We gradually increase the batch size (from size 3 in our system), to the point where the MLLM must select among up to 100 options at a time. The consequence of doing this is shown in the ``$-$ Vis. Scale.'' row in \Cref{tab:IMAGE_Ablations}. Anchor objects and interaction surfaces chosen become significantly less accurate, shown by the substantially lower energy score, plausibility, and L2 errors. In fact, removing inference-compute scaling for \selectionmethodname\ lowers performance along three of the four metrics below that of ablating low-level geometry, plausibility pruning, \textit{and} removing visual-selection. This shows the importance of 
\selectionmethodname\ to \methodname. More in the \supp.


\section{Conclusion \& Discussion}

\methodname\ is a novel approach to address the task of 3D object placement by integrating the geometric reasoning capabilities of 3D processing tools with the common-sense reasoning of MLLMs. By leveraging inference-compute scaling for the visual selection task 
through \selectionmethodname, 
\methodname\ translates language instructions into grounded 3D constraints, producing object placements that are not only geometrically feasible but also semantically plausible when considering factors like aesthetics, functionality, and accessibility. \methodname\ highlights the potential for MLLMs in 3D environments to solve spatial reasoning tasks when complemented by external geometric reasoning tools.  

\section*{Acknowledgements}
We'd like to acknowledge the SpatialVerse team at Manycore Tech Inc. for their provision of the 3D scene assets used in this project. Finally, we'd like to thank Kyle Genova and Tom Funkhouser for their feedback on the paper draft.

{
    \small
    \bibliographystyle{ieeenat_fullname}
    \bibliography{main}
}

\lstset{
  basicstyle=\ttfamily,
  columns=fullflexible,
  frame=single,
  breaklines=true,
}
\clearpage
\setcounter{page}{1}
\maketitlesupplementary


In \Cref{sec:limitations}, we discuss the limitations and potential improvements to \methodname, and comment on societal impact in \Cref{sec:impact}.

We elaborate on the designs of metrics (specifically the energy and plausibility scores introduced in the main paper) in further detail in \Cref{sec:metric_design}.

We discuss the prompts and algorithms used in constraint outline (\Cref{sec:constraint_outline_generation}), anchor object extraction (\Cref{sec:anchor_extract}), interaction surface extraction (\Cref{sec:surface_extract}) and continuous parameter estimation (\Cref{sec:param_estimation_implementation}). We also share the implementations of our constraint functions in \Cref{sec:constraint_func_implementation}.

In \Cref{sec:baseline_prompts}, we explain how we adapted Holodeck and LayoutGPT for the object placement task, and discuss the prompts used. Further details on the evaluation dataset is provided  in \Cref{sec:eval_dataset}.

In \Cref{sec:ablation_explanations}, we show qualitative consequences of ablations done in the main paper, and also do additional experiments on the performance effects of scaling \textit{down} inference compute used in \selectionmethodname, further demonstrating the benefit of scaling inference compute for the visual selection task.

In \Cref{sec:image_inputs}, we show the performance of \methodname\ on \textit{image} inputs that depict a single example of an object placement, where \methodname\ is tasked to generate similar placements. Finally, in \Cref{sec:mllm_comparison}, we show superior performance of \methodname\ over baselines (for both text and image inputs), based on comparisons done by an MLLM.


\section{System Limitations} \label{sec:limitations}

Since our method uses MLLMs in every step of the placement generation process (with the exception of Step 4 in \Cref{fig:problem_solving_process}), latency is a limitation of our approach. Our system currently takes on the range of 30 seconds to 2 minutes per object placement, depending on the number of prior objects within the scene (with more anchor object candidates, \selectionmethodname\ must select among more options, creating more calls to the MLLM) and the number of surfaces that get extracted. Additionally, much of the computation time was used for rendering object placements. For our experiments, we only used CPUs for \methodname, so it is likely possible to be sped up using GPU rendering.

\begin{figure}[tb]
    \centering
    \includegraphics[width=\columnwidth]{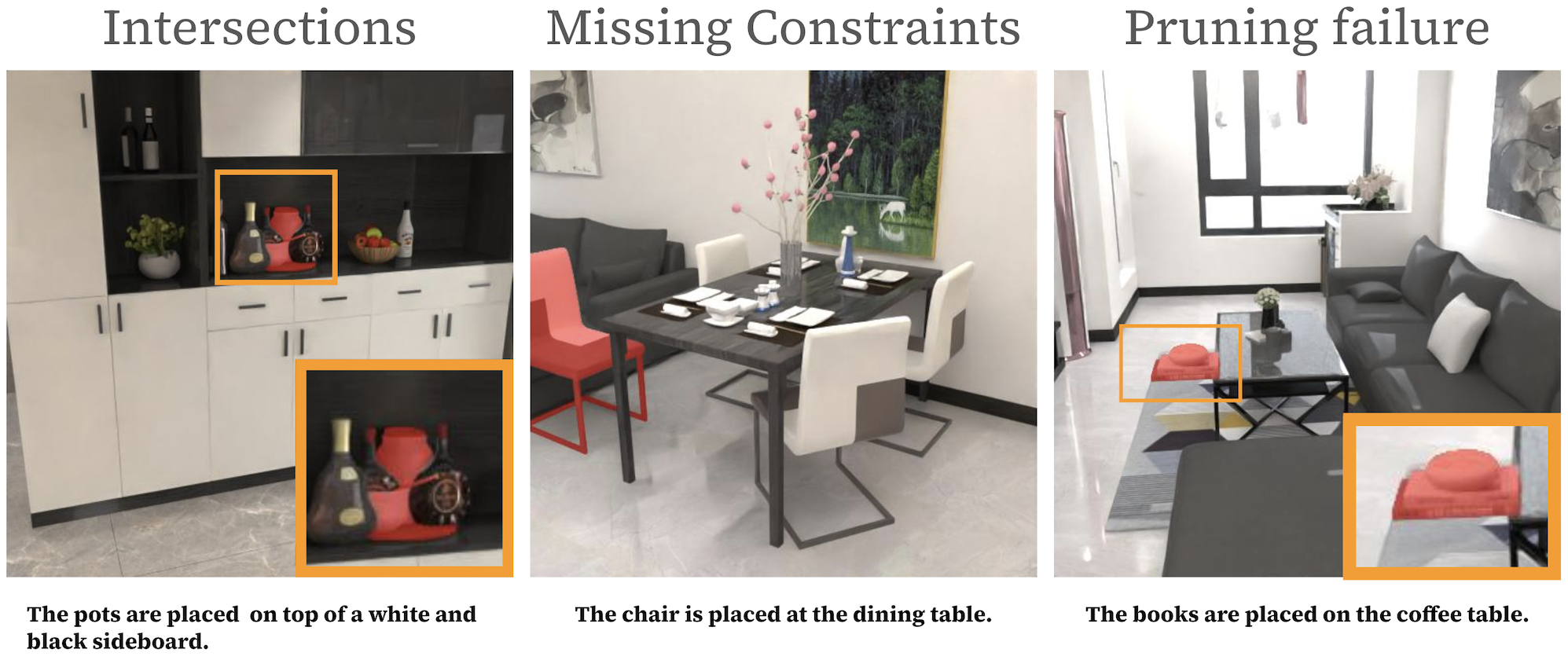}
    \caption{\textbf{Common failure modes.}  On the left, the placement of the object overlaps with preexisting objects, due to the constraint library not including a constraint to minimize intersections. In the middle, the placement of the chair was not constrained beyond contact to the ground, but additional constraints should have been generated (such as parallelism between the backs of the masked chair and the adjacent chair). On the right, the plausibility pruning step failed to remove implausible placements in the event of under-constrained placements (the bottom of the books are in contact with the table, but is overhanging), leading to a placement result that features the book floating over the edge of the table.}
    \label{fig:general_failurecases}
\end{figure}

\Cref{fig:general_failurecases} shows some qualitative examples of common failure cases for \methodname. They often come from intersections between the placed objects and preexisting objects within the environment (which can be addressed using intersection constraints), or failure to generate comprehensive sets of constraints (leading to under-constrained placements), or failure to prune the set of generated placements, often when the item being placed is small compared to the rest of the scene. In other cases, \methodname\ may choose the incorrect object for the anchor object or the incorrect surfaces of objects, a limitation inherited from existing MLLMs. As MLLMs improve, we expect these issues to be mitigated.

\begin{figure}[tb]
    \centering
    \includegraphics[width=\columnwidth]{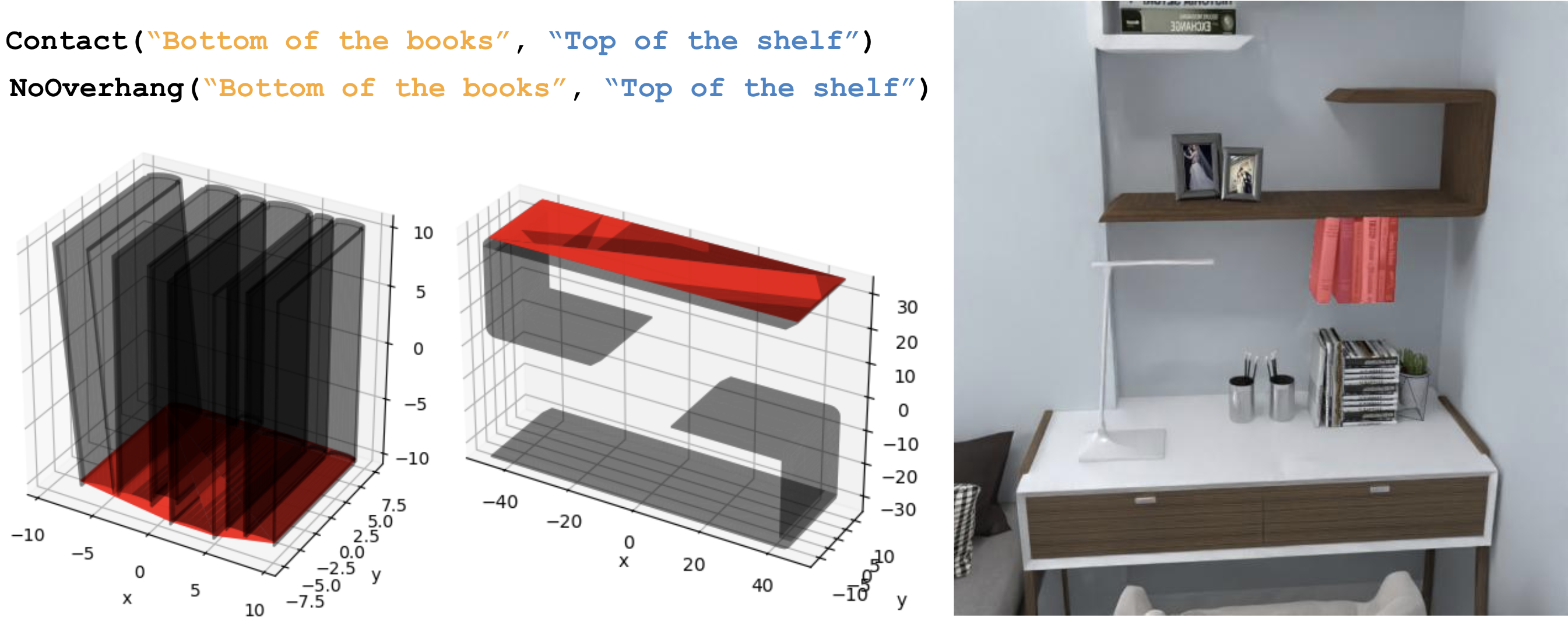}
    \caption{ \textbf{When surface extraction is done in canonical space, but the object is rotated in world space.} A failure case where a stack of books is placed \textit{under a shelf} because contact constraints were enforced for a upward-pointing (in the canonical space) surface of the shelf that \textit{has been rotated by the artist} in its final position in the world frame.}
    \label{fig:bookshelf_failurecase}
\end{figure}

Our surface extraction method extracts surfaces of anchor objects in canonical space. This assumes that surfaces of anchor objects that are pointing, say, upwards in the canonical space are \textit{also} pointing upwards in the world space once the anchor object is transformed. In rare cases within our dataset, this is not a valid assumption, the resultant optimized placements reflect this. An example can be seen in \Cref{fig:bookshelf_failurecase}.

\section{Societal Impact} \label{sec:impact}

We do not foresee any substantial negative impacts of our work, beyond the inheritance of potential bias that may already be inside the MLLMs that currently exist. We anticipate that ongoing and future efforts in reducing MLLM bias will mitigate this.


\section{Design of Metrics} \label{sec:metric_design}

Evaluating object placement is tricky. On the one hand, we have a groundtruth placement created by human artists with which we can compare generated placements, but such groundtruth accounts for only \textit{one} of \textit{many} possible object placements (\eg consider placing a cup on a table). To mitigate this, energy and plausibility scores guard against the ``precision'' and ``recall'' of the constraint functions that are generated.

Consider the simple example of placing a cup on top of a table. Our system creates constraint functions that should be \textit{minimally valued} at the groundtruth positions. This is what the \textit{energy score} communicates -- the \textit{proportion} of constraint functions constructed that are minimal ($<0.01$) when evaluated on the groundtruth placement generated by the human artist. If energy score is 1, this means that all the constraint energy functions generated for placing the cup (e.g. parallel constraints between bottom of cup and the top of the table) are ``correct'' according to the one groundtruth example. The energy score can be low when the placement is \textit{over}-constrained, or when the constraint functions have low ``precision''. 

\begin{figure*}[tb]
    \centering
    \includegraphics[width=0.8\textwidth]{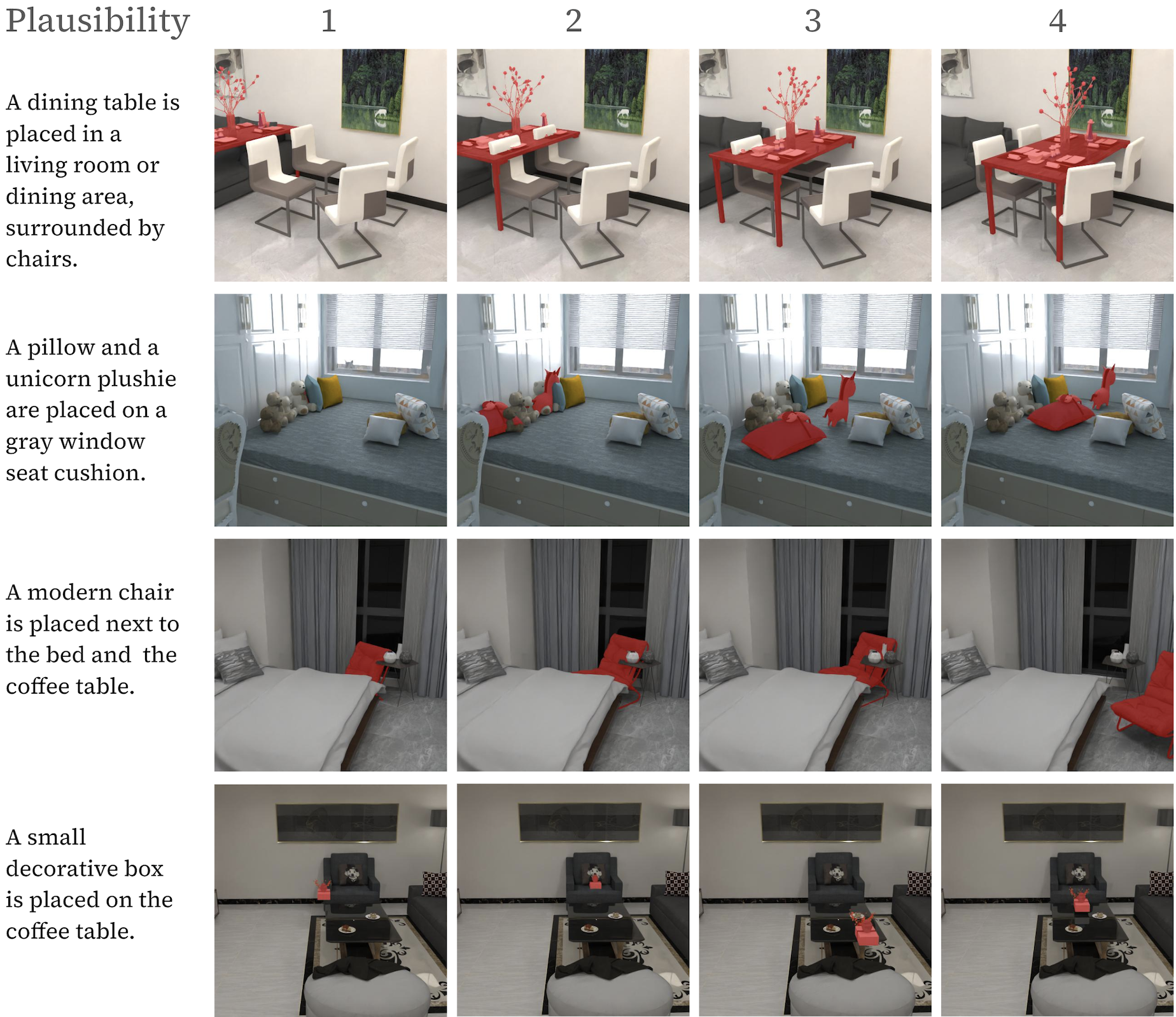}
    \caption{Examples of plausibility scores for different placements. The text prompts are shown on the left, and objects placed are shown on the right. A plausibility score of 4 is the maximum, and a 1 is the minimum. Refer to \Cref{fig:plausibility_prompt} for the definitions of these scores.}
    \label{fig:plausibility_examples}
\end{figure*}

On the other hand, what if the constraints are \textit{under}-constrained? The plausibility score seeks to measure this. Assume that the parallel constraint mentioned above was the \textit{only} one generated. As you may imagine, many solutions satisfying this constraint would render the object floating in parallel to the table, but not necessarily in contact or within the tabletop's perimeter. This would score poorly on plausibility by the MLLM, since many final renderings will display placements that are not physically realistic or disagreeing with the input text prompt (see \Cref{fig:plausibility_examples} for examples). In this case, the plausibility score will indicate that the placement is \textit{under}-constrained, or that the constraint functions have low ``recall''. The prompts (and rubric) used to evaluate the plausibility score are given in \Cref{fig:plausibility_prompt}.

Prior works like \cite{wu2024gpt} use MLLMs for evaluation, and have shown through human evaluations that MLLM evaluations are aligned with human preferences. In this paper, we observe the same for the plausibility score. As shown in \Cref{fig:plausibility_examples}, plausibility scores capture the extent to which placements are physically feasible and semantically plausible. Additionally, as noted in the main paper, when using plausibility scores to decide the better placement between two samples, they agree with humans 89.92\% of the time, when using Gemini 1.5 Pro. See \Cref {sec:human_eval} for more.


\section{Constraint Outline Generation} \label{sec:constraint_outline_generation}
We use the prompt shown in \Cref{fig:constraint_outline_generation} to query the MLLM to generate constraint outlines. As part of the prompt, we also provide the rendering of the input 3D scene (\colorbox{light-gray}{[Source Layout Rendering]}), the input text prompt (\colorbox{light-gray}{[Placement text prompt]}) and a doc string describing the constraint functions within our constraint function library (\colorbox{light-gray}{[Constraint library doc string]}). The contents of this doc string is shown in \Cref{fig:constraint_documentation}.

\section{Extraction of Anchor Objects}\label{sec:anchor_extract}

\begin{figure}[tb]
    \centering
    \includegraphics[width=\columnwidth ]{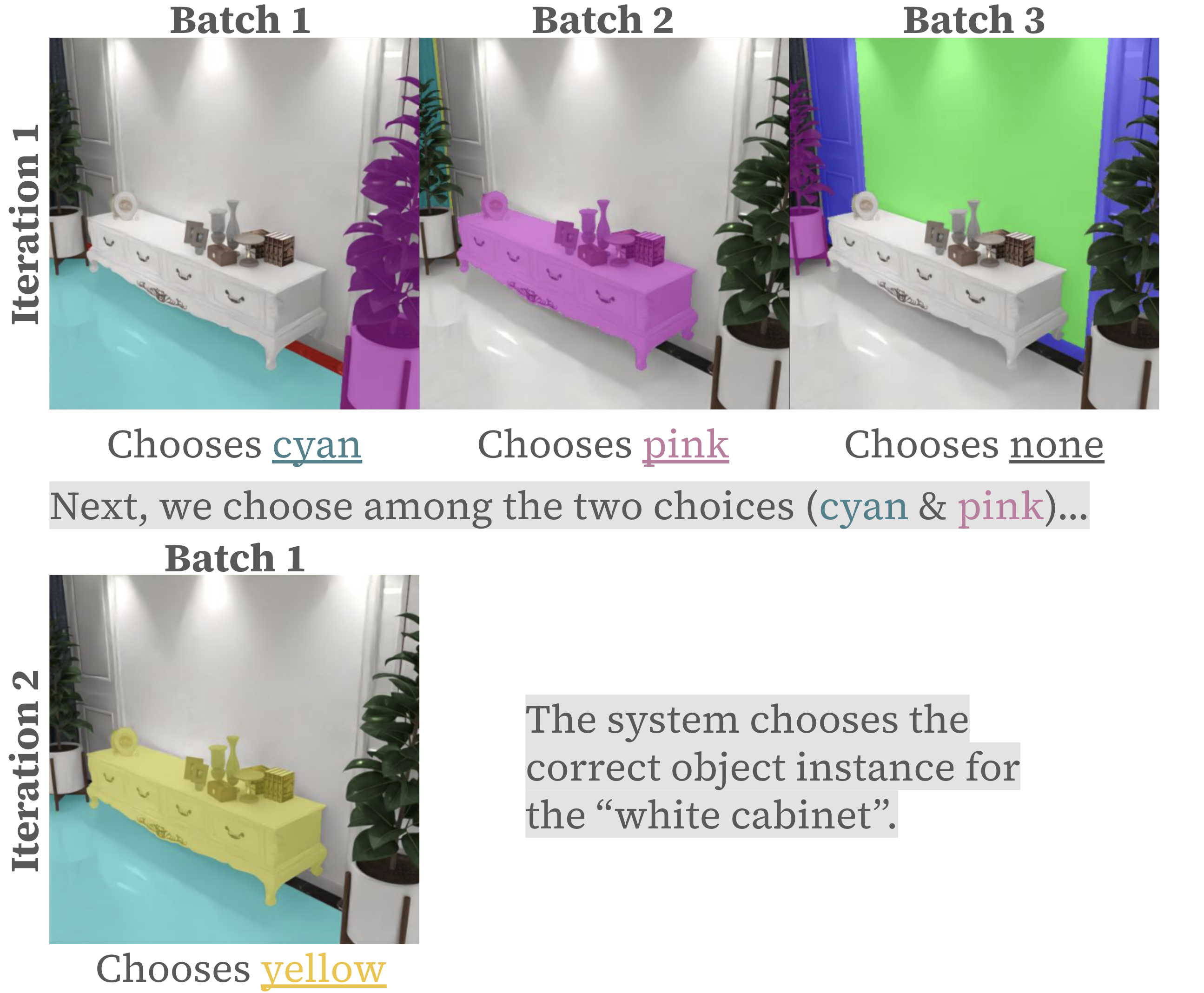}
    \caption{The batched visual selection process for a scene with only a few items. Here, the MLLM is tasked to find the anchor object corresponding to ``the white cabinet'' from a constraint outline generated. Each batch shows 3 options rendered in different colors (for batch size $= 3$), and the MLLM chooses object instances that best match the description by indicating the color of the mask in each round. The chosen instances across each batch are merged and the process is repeated until only one object instance is chosen. This is done using the prompt shown in \Cref{fig:extract_anchor_object}.}
    \label{fig:batched_visual_selection_process}
\end{figure}

We use \selectionmethodname\ to select for object instances that match language descriptions of the anchor objects from the constraint outline. Beyond the procedure outlined in \Cref{alg:visual_selection_algo}, we use the prompts shown in \Cref{fig:extract_anchor_object} to select among every batch. The result of doing so is shown in \Cref{fig:batched_visual_selection_process}. Their image segmentation masks are derived from the USD rendering process of the input 3D scene.

\section{Extraction of Surfaces} \label{sec:surface_extract}

To execute the surface extraction steps in \Cref{alg:constraint_construction}, we will first describe how \textbf{DirExtr()} works, then elaborate on the geometric processing underlying \textbf{SurfExtr()}. 

In \textbf{DirExtr()} the MLLM is prompted to generate surface normals that match the language descriptions of the surfaces that best match the constraint outline descriptions of surfaces that participate in the constraint. For instance, if the transformable object should be sitting on the seat of the chair, the surface that should be extracted from the chair should be  pointing upwards (\ie the seat). This is done through the prompt shown in \Cref{fig:extract_surfaces}. Note that in our experiments, we provide the MLLM with the 6 major directions (left, right, front, back, up, down) that surfaces can point, but our method can also work with more directions by updating the prompt accordingly. After this is done for the anchor and transformable objects, geometric processing algorithms are then called to extract the surfaces that point in these directions.

To this end, we first filter the faces of the object mesh to a subset that have face normals within some threshold level of cosine similarity with the unit vectors corresponding to the extraction direction generated in the previous step. For the faces that have similar face normals to the extraction direction, we project the center of these faces along the desired surface normal, then cluster them using DBSCAN. This allows us to find different sets of faces that lie along a similar ``level'' along the desired surface normal. Faces in each set are then projected to the same level before a convex hull is fitted onto that group, which extracts a flat convex hull tightly circumscribing each set. Each of these planar convex hulls is an interaction surface candidate, and has a surface normal equal to the extraction direction. 

This process leads to many candidate interaction surfaces and, depending on the level of geometric complexity of assets, may be prohibitively cumbersome/expensive/difficult to filter down to one using \selectionmethodname. As such, we merge interaction surfaces that are close to each other within a certain distance threshold by keeping the interaction surfaces that have a larger area. Additionally, we recognize that many constraints can be adequately expressed using bounding box constraints, so we append the bounding box surface aligned with the extraction direction to the set of candidate interaction surfaces.

The process of choosing the correct interaction surfaces for the anchor and transformable objects is very similar to that of choosing the anchor object (See \Cref{sec:anchor_extract}). We render the surfaces overlaid on top of the original object mesh using \verb|matplotlib|, taking care to show no greater than 3 candidates for every batch. The selection is done according to the prompt shown in \Cref{fig:choose_surface}.

\begin{figure*}[tb]
    \centering
    \includegraphics[width=0.9\textwidth]{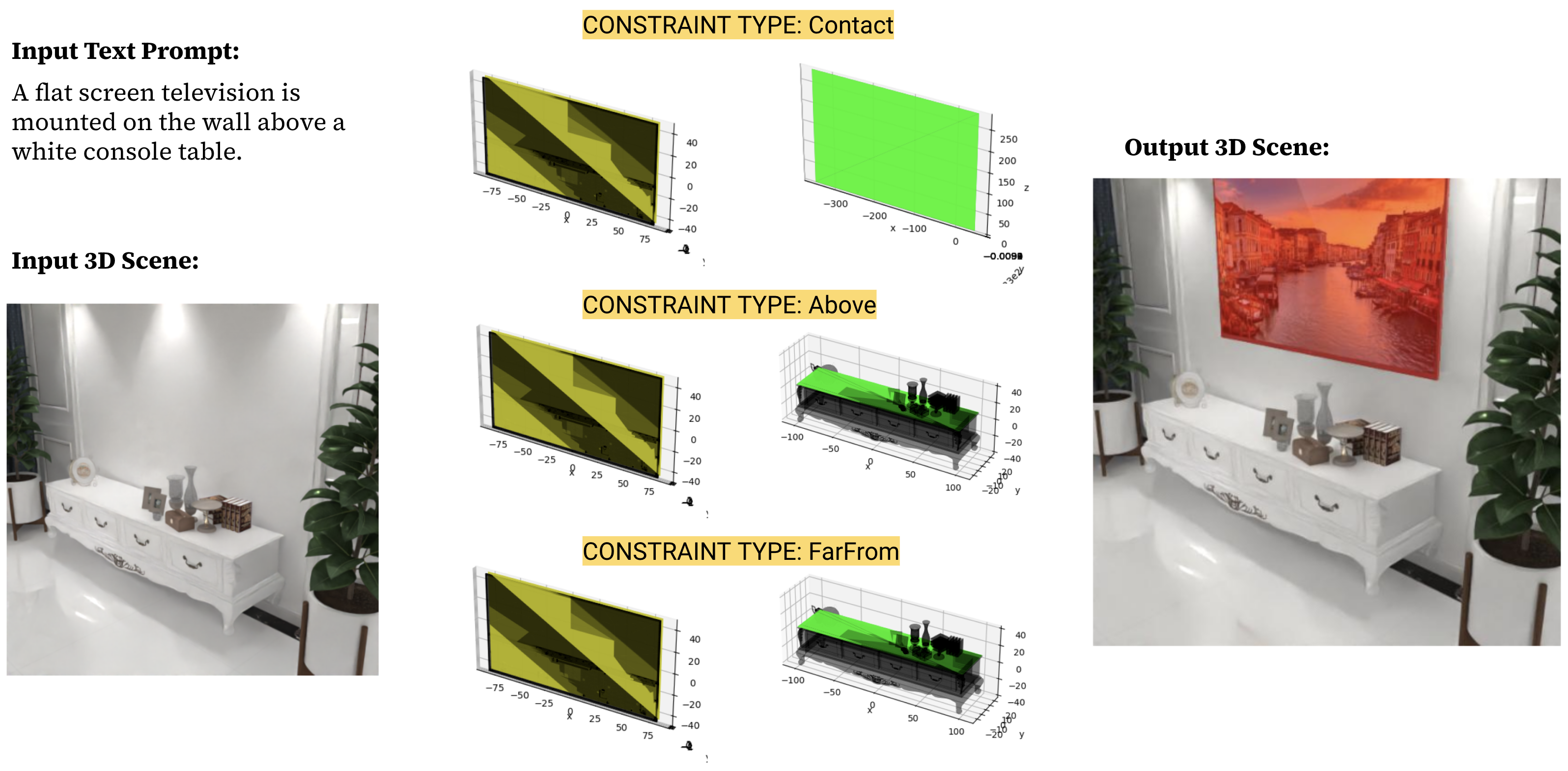}
    \caption{The constraints and interaction surfaces generated for the task of mounting a TV. For clarification, the contact constraint is enforced between the back of the TV and the wall (visualized as a plane). Note that there are multiple wall meshes within this scene (for instance, see \Cref{fig:batched_visual_selection_process} -- blue in Batch 3 and cyan in Batch 2 are both alternatives), and that \selectionmethodname\ chooses the correct one }
    \label{fig:planes_ex1}
\end{figure*}

\begin{figure*}[tb]
    \centering
    \includegraphics[width=0.9\textwidth]{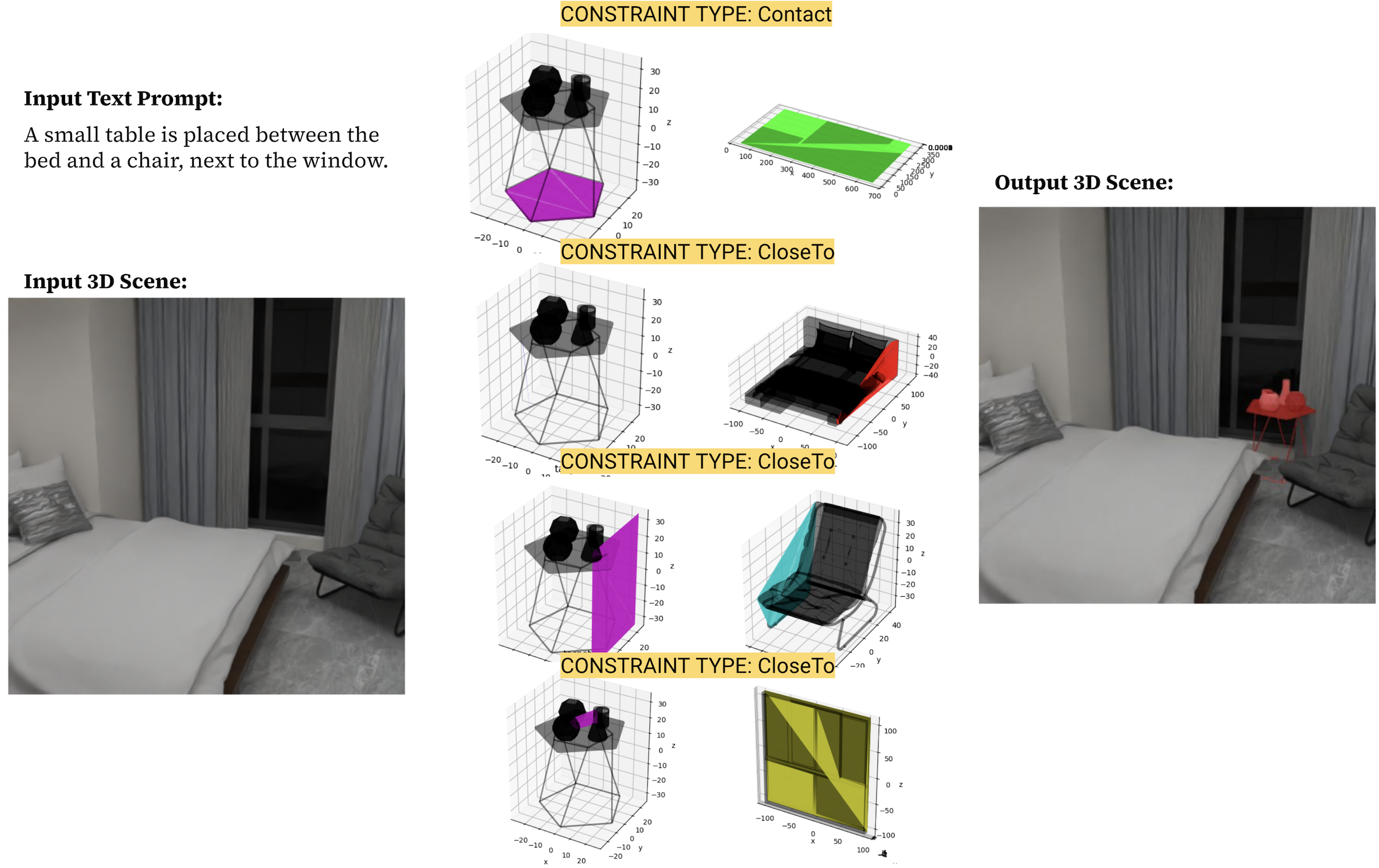}
    \caption{To place the small table into the  scene, the constraints generated first identifies a Contact constraint between the bottom of the table and the floor, then uses various CloseTo constraints to capture its rough position in the room - the final position generated is close  to all 3 surfaces chosen.}
    \label{fig:planes_ex2}
\end{figure*}

\begin{figure*}[tb]
    \centering
    \includegraphics[width=0.9\textwidth]{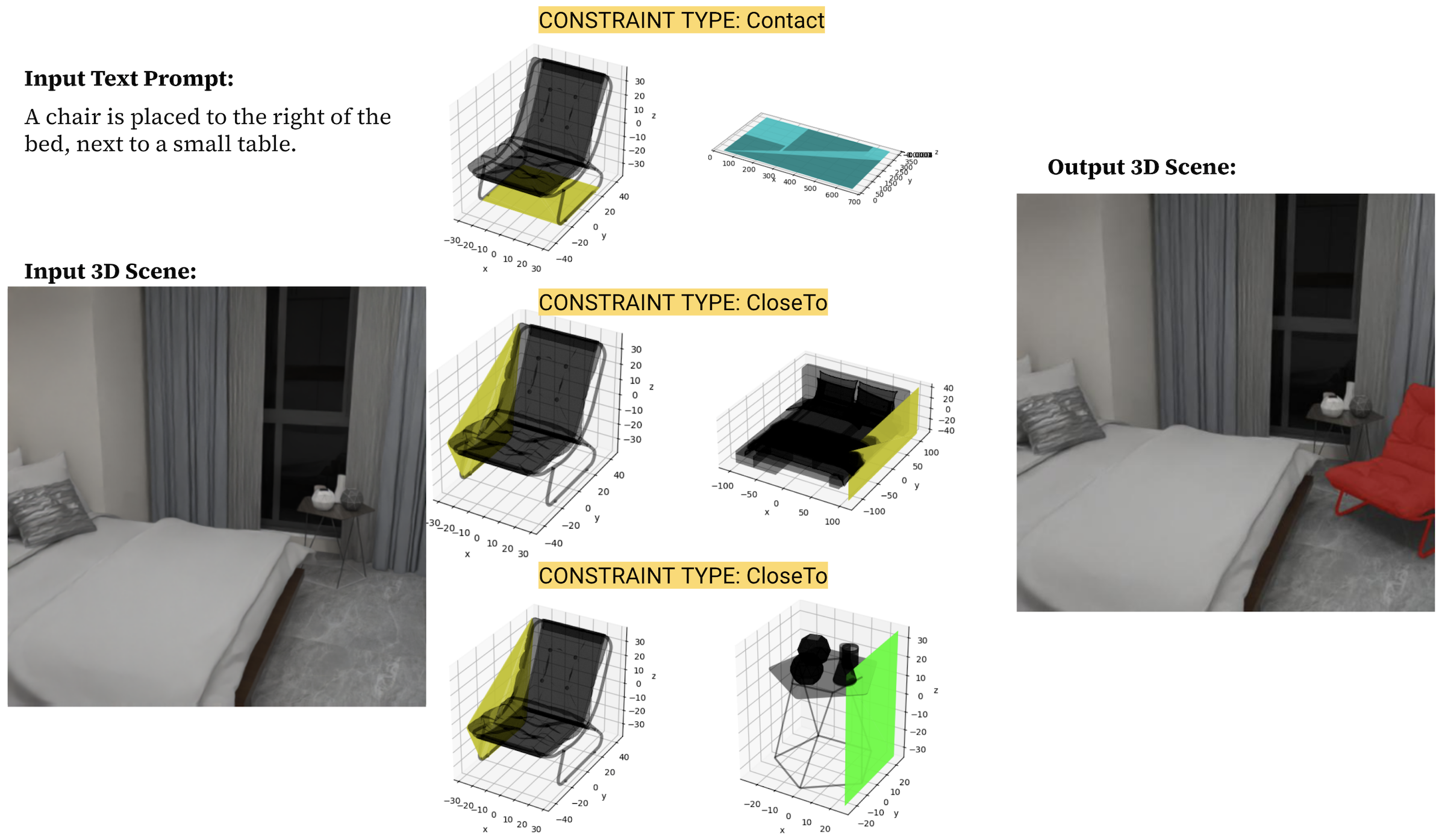}
    \caption{Similar to \Cref{fig:planes_ex3}, \methodname\ enforces a contact constraint with the floor, then uses CloseTo constraints to restrict plausible placements.}
    \label{fig:planes_ex3}
\end{figure*}

\begin{figure*}[tb]
    \centering
    \includegraphics[width=0.9\textwidth]{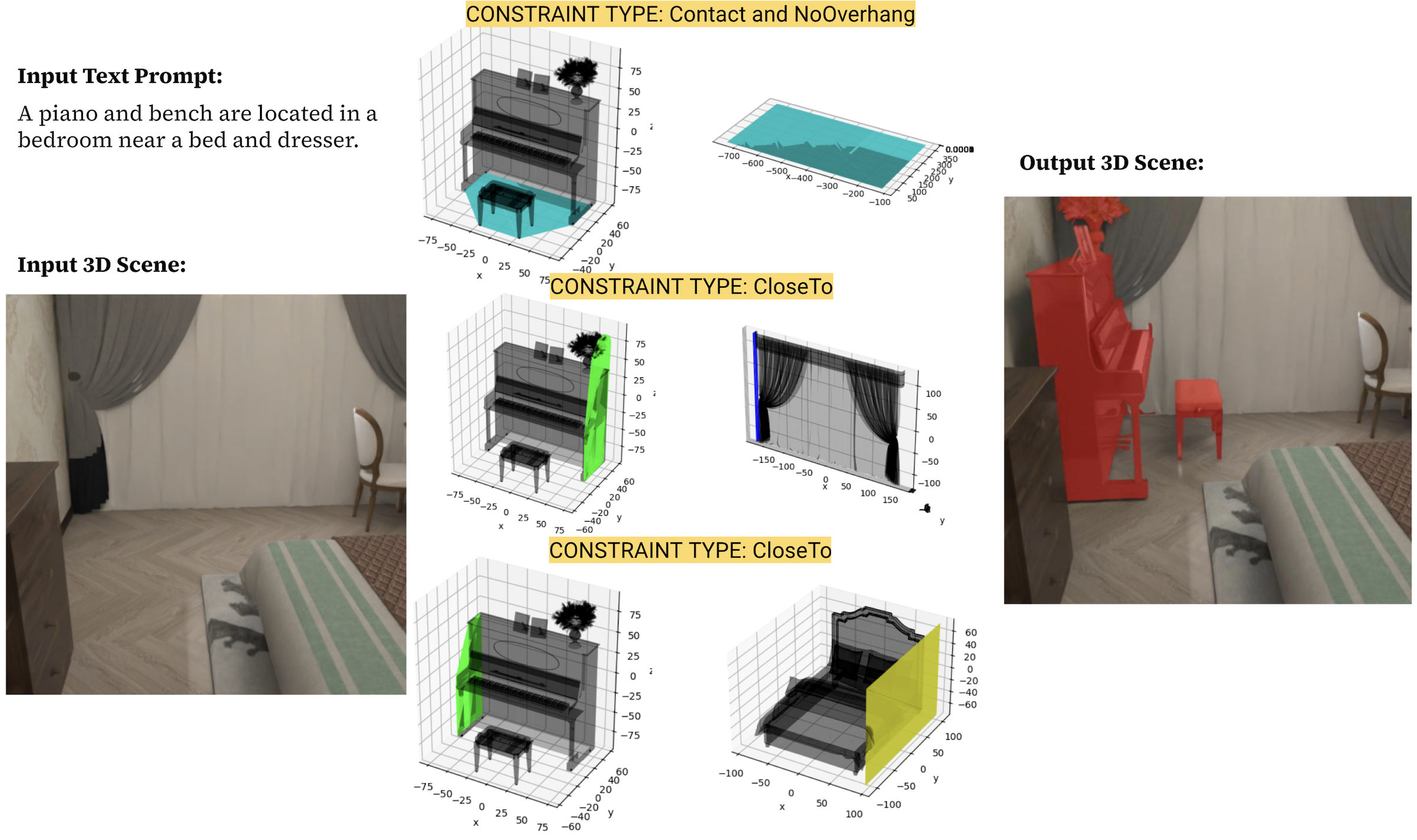}
    \caption{Placing a piano into the room. \methodname\ successfully discerns that the left and the right of the piano must be close to different things, and that the right side should be closer to the left of the curtains, and the left should be closer to the bed.}
    \label{fig:planes_ex4}
\end{figure*}

\begin{figure*}[tb]
    \centering
    \includegraphics[width=0.9\textwidth]{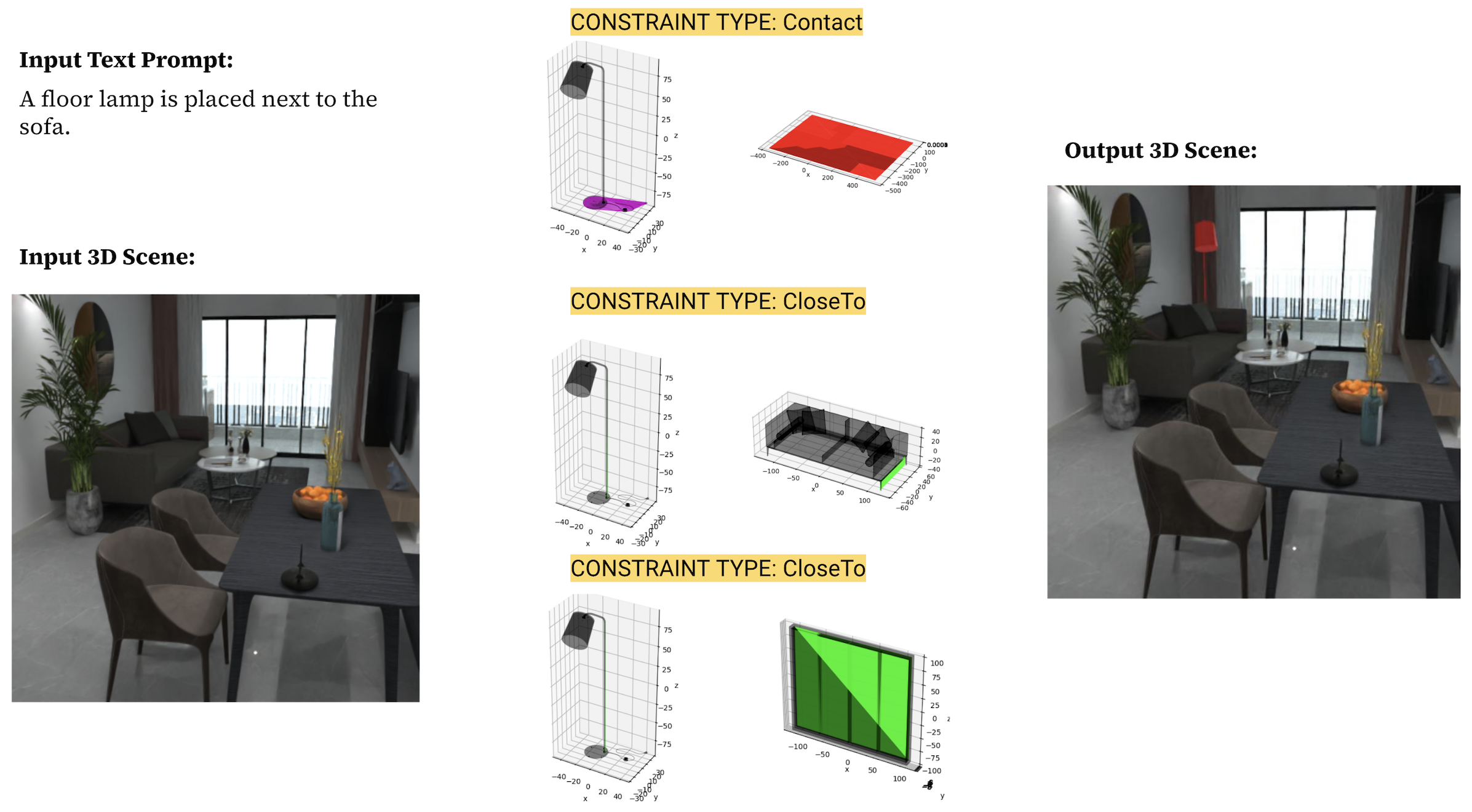}
    \caption{Placing the lamp in the living room is done by locating near-by objects and enforcing CloseTo constraints.}
    \label{fig:planes_ex5}
\end{figure*}

\begin{figure*}[tb]
    \centering
    \includegraphics[width=\textwidth]{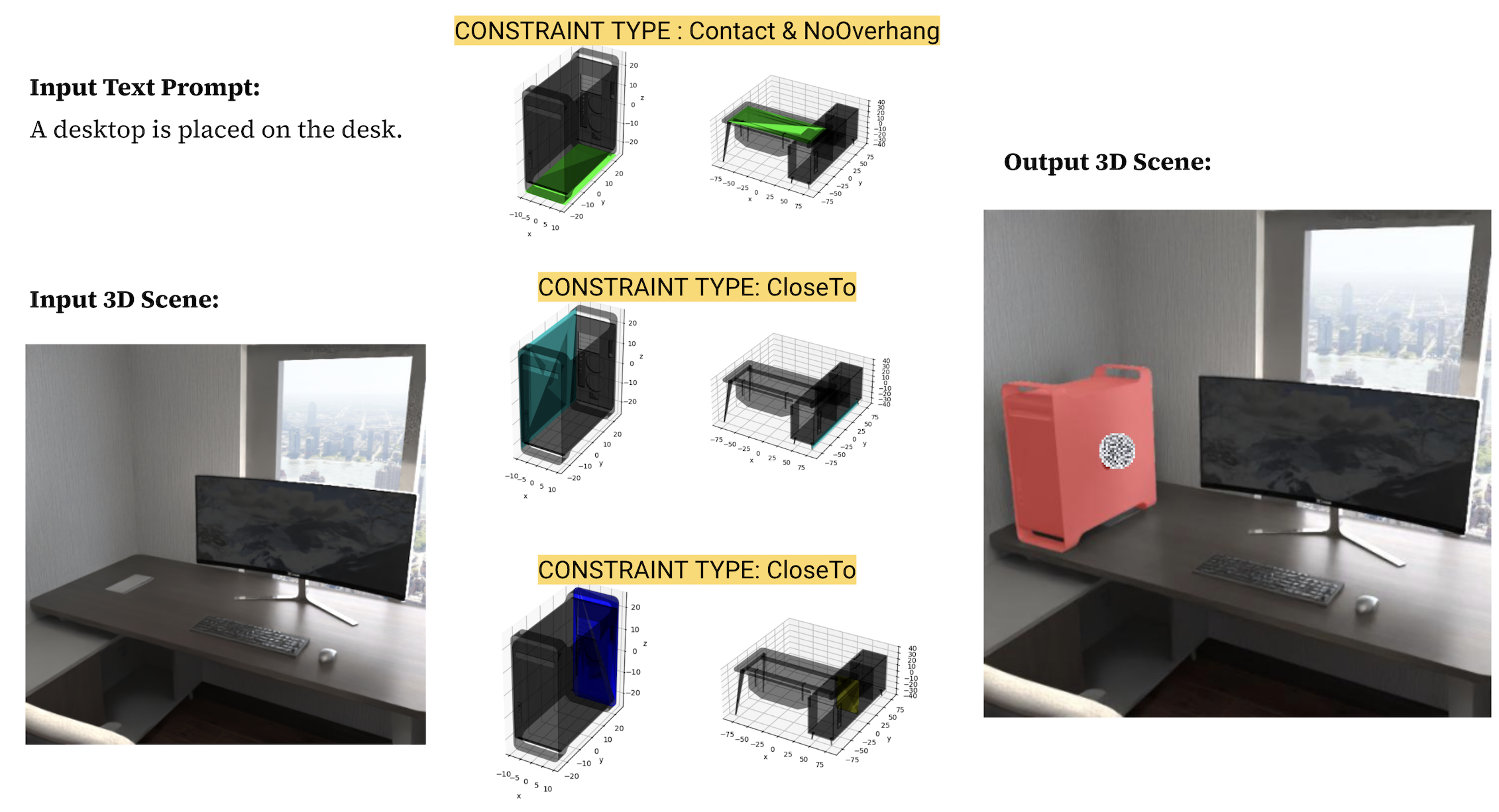}
    \caption{The desk is L-shaped, meaning that in order to insert place the desktop, it must correctly extract the top of the desk, and not just use the top face of the desk's bounding-box. We see here that it enforces Contact and NoOverhang constraints to the table, guides the placement according to other CloseTo constraints.}
    \label{fig:planes_ex6}
\end{figure*}

\Cref{fig:planes_ex1}, \Cref{fig:planes_ex2}, \Cref{fig:planes_ex3}, \Cref{fig:planes_ex4}, \Cref{fig:planes_ex5}, \Cref{fig:planes_ex6} show examples of the constraints constructed by \methodname\ for different placement tasks using the surfaces extracted by this approach.

\section{Parameter estimation} \label{sec:param_estimation_implementation}

Once interaction surfaces of the anchor and transformable objects are extracted, we prompt the MLLM with renderings of the scene and \verb|matplotlib| renderings of the interaction surfaces, according to the prompt shown in \Cref{fig:estimate_parameters}. A documentation of the meaning of the continuous parameters of each constraint function is also provided. In our setting, only two of the constraint functions have continuous parameters (\verb|CloseTo| and \verb|FarFrom|); namely, maximum/minimum distances between the two interaction surfaces.

\section{Constraint functions} \label{sec:constraint_func_implementation}

The constraint functions outlined in \Cref{sec:constraint_lib} are implemented as binary functions that evaluate to $0$ when a geometric relationship is satisfied between two interaction surfaces, \verb|p1| and \verb|p2|. Below, we describe the implementations of each:
\begin{enumerate}
\item \verb|Parallel(p1, p2)| returns $$\min(|1 - n_1^Tn_2 |, |-1 - n_1^Tn_2|)$$ where $n_1$ is the surface normal of \verb|p1| and $n_2$ that of \verb|p2|. This function is minimal when either the surface normals are aligned, or pointing in parallel but opposite directions.
\item \verb|CloseTo(p1, p2, dist)| returns $$k \max(d(\verb|p1|, \verb|p2|) - \verb|dist|, 0)$$ for some scaling constant $k$ (which we set to 0.1) and distance function $d$ between two interaction surfaces. This function is minimal when $d(\verb|p1|, \verb|p2|)$ is \textit{smaller} than \verb|dist|.
\item \verb|InFrontPlane(p1, p2)| first finds the pairwise vector differences between the vertices of $\verb|p1|$ and $\verb|p2|$, then evaluates the dot product between each vector difference and the surface normal of $\verb|p1|$, $n_1$. The return value of this function is $$\min(0, \max(\{ - \frac{v_{ij}}{|v_{ij}|}^T n_1 \}_{ij}))$$ for all vector differences $v_{ij}$ between the $i$th vertex in $\verb|p1|$ and the $j$th vertex in $\verb|p2|$.
\item \verb|Contact(p1, p2)| is $$\verb|InFrontPlane(p1,p2)| + \verb|InFrontPlane(p2,p1)|$$ which is minimal when \verb|p1| and \verb|p2| are either in contact with each other or coplanar.

\item \verb|NoOverhang(p1, p2)| is more involved. Let $p_1$ be the set of vertices of \verb|p1| (with N vertices and 3 coordinates, $p_1 \in \mathbb R^{N \times 3}$) and $p_2$ be that of \verb|p2|. Let $n_2$ be the normal vector of \verb|p2|. We then first project $p_1$ onto \verb|p2|, $$p_1 |_{\verb|p2|} = p_1 - ((p_1 - o)\cdot n_2)n_2$$ for some arbitrary vertex $o$ from $p_2$. Then, we sample $1000$ points from the region bound by $p_1 |_{\verb|p2|}$, which we call $q \in \mathbb R ^{1000 \time 3}$. We can then calculate whether $q$ is contained within the bounds of $p_2$. The final output value of this function is defined as $$ 1 - \frac {1}{1000} \sum_{i=1}^{1000} \mathbb I_{inside} (q_i)  $$
where $\mathbb I_{inside}$ is an indicator function indicating whether $q_i$ lies on the inside of \verb|p2|. The function is minimal when the projection of \verb|p1| onto \verb|p2| is entirely contained within \verb|p2| (\ie $\forall i, \mathbb I_{inside}(q_i) = 1$).
\end{enumerate}

\section{Prompts and Constraint Functions for Holodeck and LayoutGPT} \label{sec:baseline_prompts}

For the Holodeck and LayoutGPT baselines in our experiments, we use the prompts shown in \Cref{fig:baseline_holodeck_prompt1}  and \Cref{fig:baseline_holodeck_prompt2} for Holodeck, and \Cref{fig:baseline_layoutgpt_prompt} for LayoutGPT. Note that we use the prompts found in the implementations released on github 
\footnote{\href{https://github.com/allenai/Holodeck/blob/main/ai2holodeck/generation/prompts.py}{Official Holodeck prompts} and \href{https://github.com/weixi-feng/LayoutGPT/blob/master/run_layoutgpt_3d.py}{official LayoutGPT prompts}},
modified only to provide additional information about the objects that already exist within the scene. To do this, we use Gemini to caption object renderings of assets found within the scene, and provide these captions to both methods via the prompts. 

For LayoutGPT, we provide the caption of each object alongside their bounding box information (length, width, height, left, top, depth, orientation) as part of the prompt. These can be derived from their local-to-global transformation matrices, as well as the length, width, height of their bounding boxes in canonical space.

For Holodeck, these captions (\eg ``a brown curtain'') are provided alongside object ID's (\eg \verb|object-34|), indicated by \colorbox{light-gray}{[Descriptions and labels of preexisting objects in the scene]} in \Cref{fig:baseline_holodeck_prompt2}. Holodeck can then reference the object ID's in the construction of the bounding box constraints. For the bounding box constraints used by Holodeck, we implement bounding box constraints as binary constraint functions, similar to those in \Cref{sec:constraint_lib}, with the big difference being that they operate on \textit{bounding boxes} instead of \textit{interaction surfaces}. The Holodeck baseline has access to a bounding box constraint library composed of the following constraints: (1) \verb|FaceTo| (that the front face of the object bounding box faces the center of another bound box), (2) \verb|near| (that objects are closer than 150 cm from each other and further than 50 cm away) (3) \verb|far| (object are further than 150 cm away from each other) (4) \verb|infront|, (5) \verb|sideof|, (6) \verb|leftof|, (7) \verb|rightof|, (8) \verb|behind|, (9) \verb|ontop|, (10) \verb|centeraligned_front|, (11) \verb|centeraligned_side|. These are all bounding box constraints originally used in the Holodeck implementation. For comparisons between Holodeck and \methodname, we use the same constraint solver (with the same parameters) to solve constraints created by both methods.

As mentioned in the paper, experiments on \methodname, LayoutGPT, Holodeck all use the same MLLM (Gemini 1.5Pro) for a fair comparison.

\section{Evaluation Dataset} \label{sec:eval_dataset}

\begin{figure}[tb]
    \centering
    \includegraphics[width=\columnwidth ]{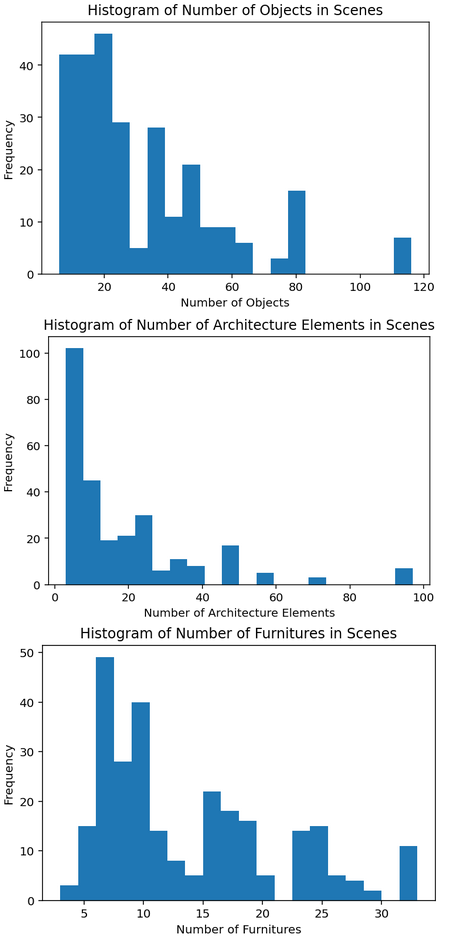}
    \caption{Distributions of the number of objects (furniture and architectural elements) within the placement tasks used for evaluation.}
    \label{fig:histogram_count_per_scene}
\end{figure}

All scenes used for evaluation of our method and the baselines are in Universal Scene Descriptor (USD) format, which contains meshes of all objects, architectural elements and photorealistic materials. When choosing the transformable objects for each placement task, our goal was to choose objects for which the correctness of its final position depends on the successful identification of ``anchor'' objects and the relevant constraints. As such, the transformable objects selected for our evaluation placement tasks are often furniture pieces (chairs, tables, refrigerators ...etc) and decorative items (books, picture-frames, wall art...etc). \Cref{fig:histogram_count_per_scene} shows the distribution of the number of possible ``anchor'' objects in each of the placement tasks composed of architectural elements and furniture/household objects. This motivates the need for \selectionmethodname\ to make the visual selection task easier by breaking down the decision process into multiple stages.

\section{Ablations}  \label{sec:ablation_explanations}
\subsection{Qualitative Examples of Ablations} 

\begin{figure}[tb]
    \centering
    \includegraphics[width=\columnwidth]{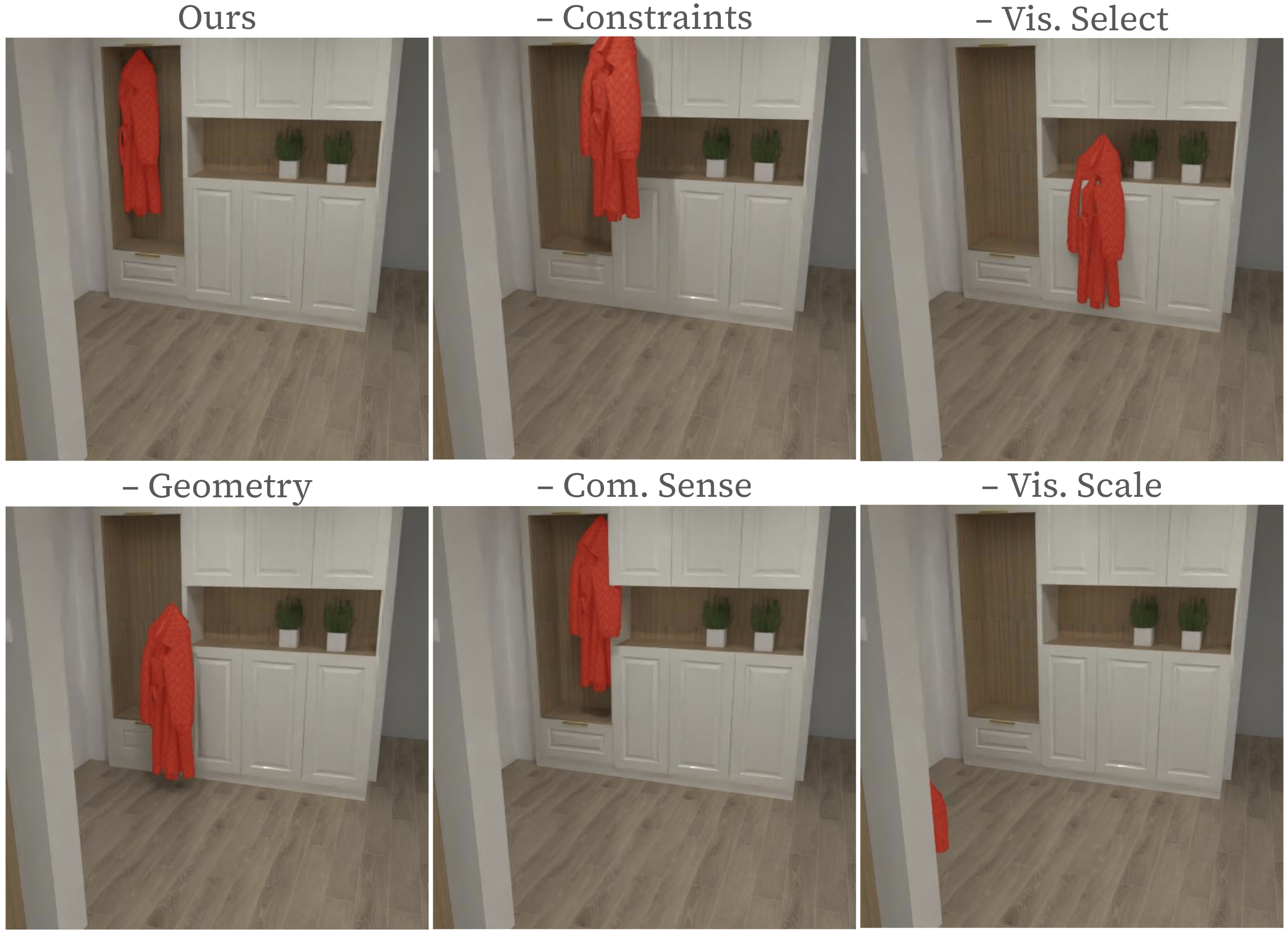}
    \caption{An example of the effects of ablations in \Cref{tab:IMAGE_Ablations} on the placements. In this example, \methodname\ is tasked to place the coat into the closet. }
    \label{fig:ablation_comparisons}
\end{figure}

\begin{figure}[tb]
    \centering
    \includegraphics[width=\columnwidth]{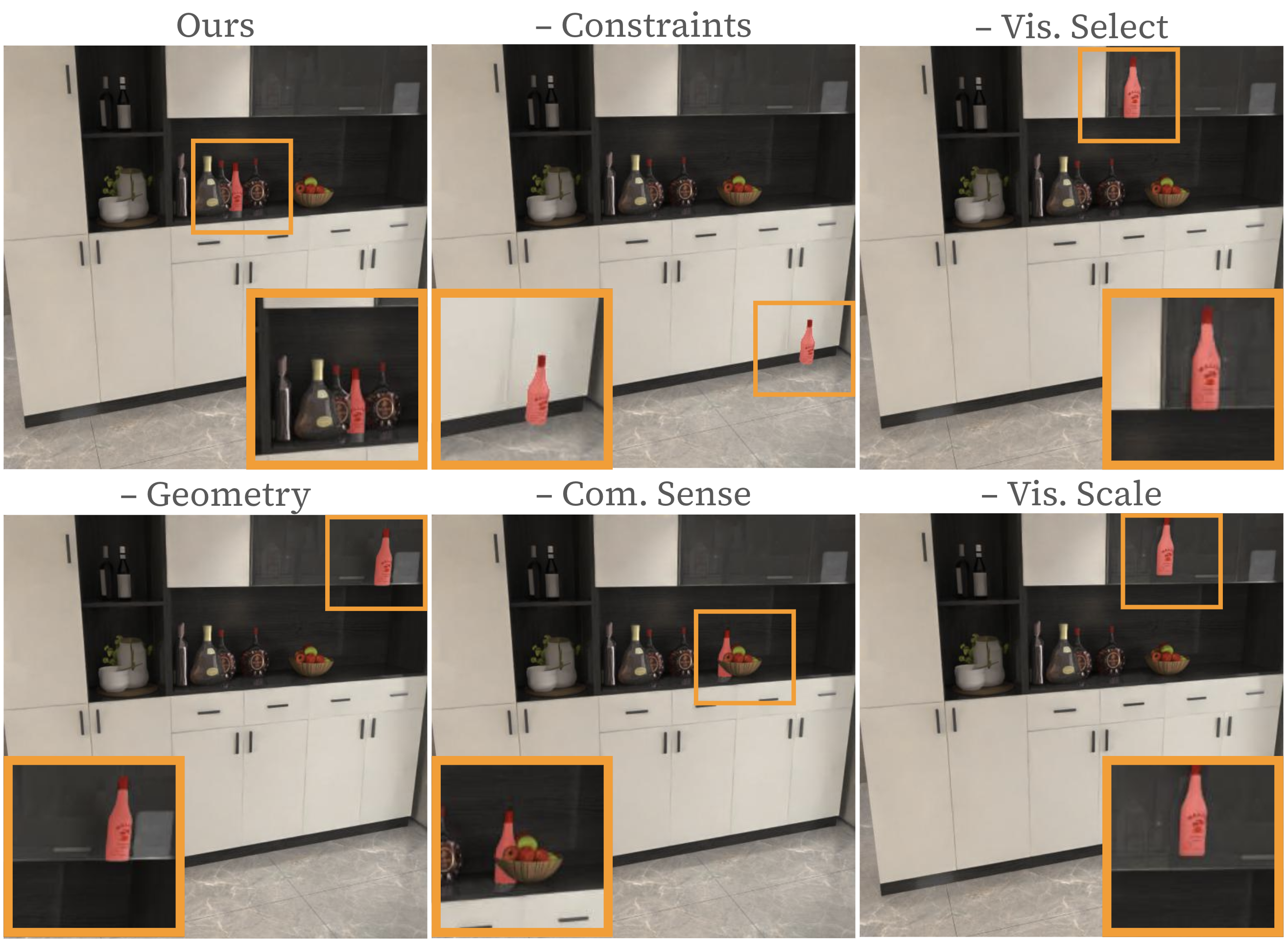}
    \caption{An example of the effects of ablations in \Cref{tab:IMAGE_Ablations} on the placements. In this example, \methodname\ is tasked to place the bottle on the cabinet.}
    \label{fig:ablation_comparisons2}
\end{figure}

The ablations tested in \Cref{tab:IMAGE_Ablations} have qualitative consequences on the placements that get generated. \Cref{fig:ablation_comparisons} and \Cref{fig:ablation_comparisons2} show this for two placement tasks. Note that in both cases, the transformable object must be placed into a shelf-like object, and that it's crucial to have access to the low-level geometry, which bounding box representation do not provide (``$-$ Geometry''). In both cases, purely using the MLLM to choose among randomly generated placements leads to suboptimal placements, oftentimes creating final placements that feature the object floating in air (\eg ``$-$ Constraints''  in \Cref{fig:ablation_comparisons2} shows the bottle floating slightly above the ground). 
We can also see that plausibility pruning tends to get rid of implausible overlaps  that may happen when placing the object according to raw geometric constraints -- in both \Cref{fig:ablation_comparisons} and \Cref{fig:ablation_comparisons2}, removing plausibility pruning (``$-$ Com. Sense'') leads to final placements that overlap with assets already in the scene. Finally, removing the ability to visually select anchor objects (as opposed to selecting anchor objects based on text annotations) and removing the ability to scale inference compute for \selectionmethodname\ both lead to incorrect placements, due to the wrong anchor object/surfaces being selected for the constraints.

\subsection{Inference Compute Scaling for \selectionmethodname} \label{sec:inference_scaling_batchsize}

\begin{figure}[tb]
    \centering
    \includegraphics[width=\columnwidth]{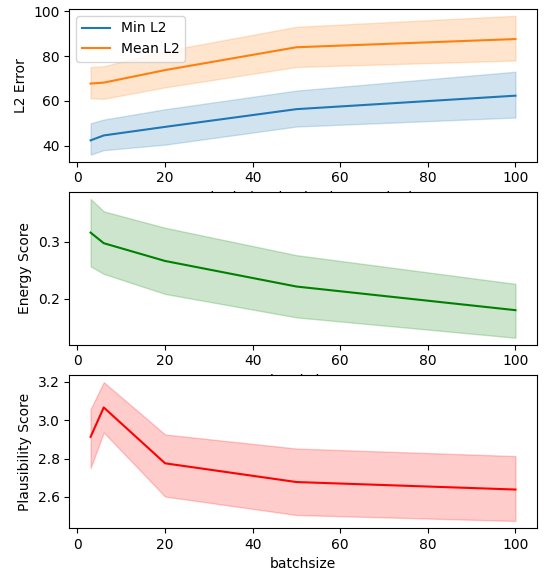}
    \caption{Increasing the batch size within \selectionmethodname\ leads to lower performance on placement tasks.}
    \label{fig:scaling_batches}
    \vspace{-1em}
\end{figure}

\Cref{fig:scaling_batches} displays the trends on the performance metrics as we increase the batch size (lower the level of inference compute) used by \selectionmethodname. This means that for the selection process of anchor objects and interaction surfaces, an MLLM must choose among larger sets of options at a time. We can observe a downward trend in plausibility and energy scores (due to incorrectly selected object instances and and interaction surfaces), and also an upward trend in both mean and minimum L2 errors, suggesting that the resultant placements become further away from the groundtruth as MLLMs are prompted to choose among more and more visual options at a time. Our default settings uses a batch size of 3, and for \Cref{fig:scaling_batches}, we increase the batch size to 6, 20, 50 and 100.

\section{Performance on Image Inputs}\label{sec:image_inputs}

\begin{figure*}[tb]
    \centering
    \includegraphics[width=0.9\textwidth]{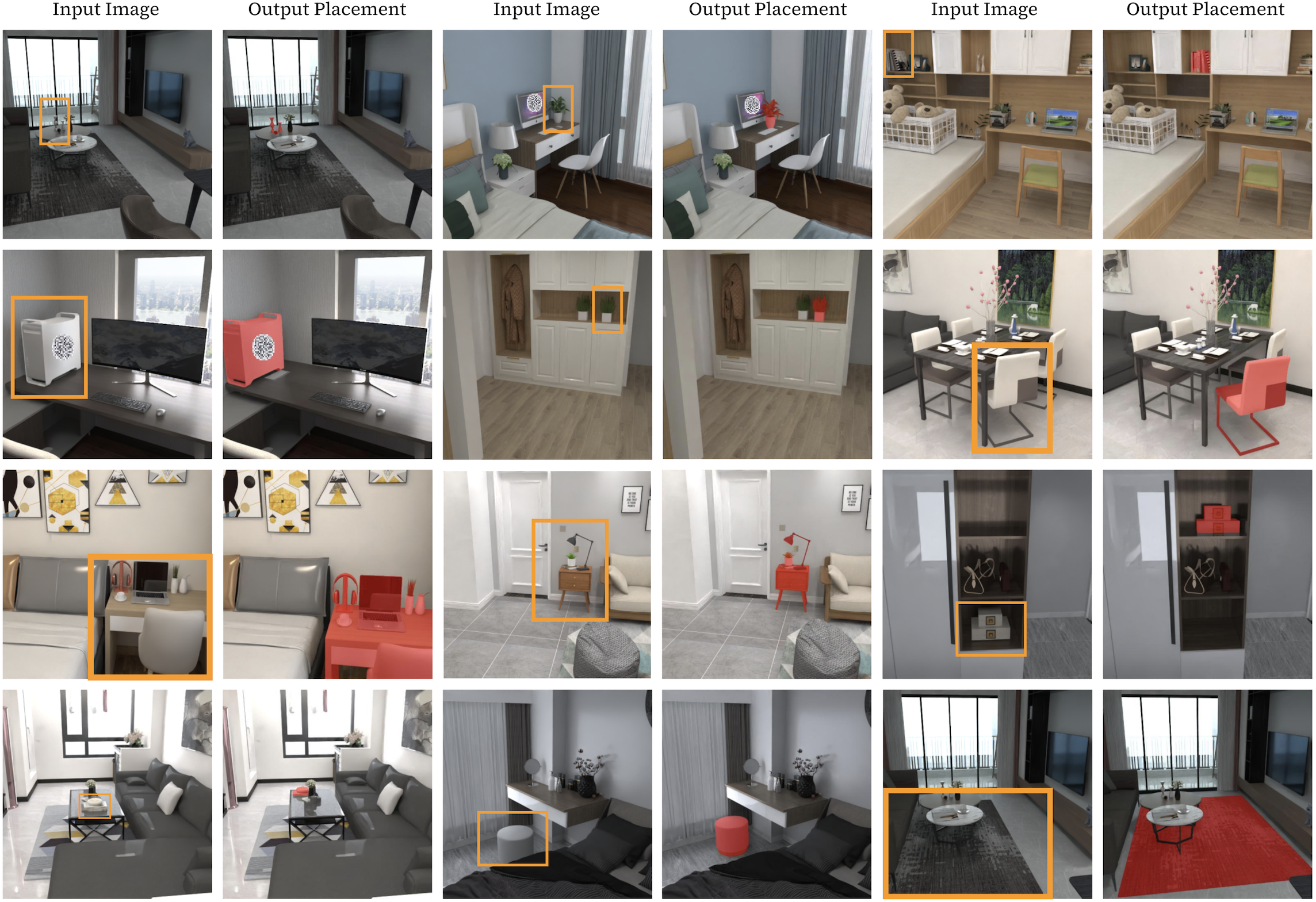}
    \caption{Qualitative results of object placement when \methodname\ is given \textit{image} inputs of placement examples. Note how generated placements follow the semantics of object placements shown in the  input image to varying degrees, but can vary in their final positions.}
    \label{fig:image_input_qualitative}
\end{figure*}


\begin{table}[tb]

    \centering
    \begingroup
    \setlength{\tabcolsep}{5pt} 
    \renewcommand{\arraystretch}{1.} 
    \begin{tabular}{@{\hskip 0.5em}rcccc@{\hskip 0.5em}}
    \toprule
    Metric & LayoutGPT & Holodeck & Ours \\
    \midrule 
Min L2 error cm ($\downarrow$) & 126.19 & 91.25  & \textbf{43.69}  \\
Mean l2 error cm ($\downarrow$) & 166.11 & 136.51 & \textbf{68.84}  \\
Visibility score ($\uparrow$) & 0.69   & 0.59   & \textbf{0.88}   \\
Plausibility score ($\uparrow$) & 2.31   & 1.99   & \textbf{2.92}   \\
Energy score ($\uparrow$)  & -- & 0.17   & \textbf{0.38}   \\
    \bottomrule
    \end{tabular}
    \caption{Comparison with Baseline for \textit{image inputs}}
    \label{tab:image_baseline_comparisons}
    \endgroup
\end{table}
Since \methodname\ uses an MLLM for constraint outline generation, an additional input modality that we can demonstrate -- besides language annotations -- is an image \textit{example}. Given an image showing one possible placement  of the  object, \methodname\ can generate variations of placements that are \textit{semantically similar} in the sense that output placements capture the underlying constraints and placement considerations of the image example. \Cref{tab:image_baseline_comparisons} shows that our method also outperforms the baselines across the metrics on this task. \Cref{fig:image_input_qualitative} shows object placements generated from our system when given different placement examples. For this experiment, the prompts of our method, Holodeck, and LayoutGPT are changed accordingly to insert the image example 
instead of a text prompt, as done for the  experiments in the main paper. 

\section{Using MLLMs to Compare \methodname\ to Baselines} \label{sec:mllm_comparison}
In addition to the human evaluations that indicate \methodname's superior performance, we can also use MLLMs to do pairwise comparisons, by giving it shuffled pairs of renderings (image A and B) generated by our method and the two baselines. The objective (text input or image input) is also provided to the MLLM. The prompt used is shown below:
\begin{lstlisting}
Between Image A and Image B, which is a better match to the objective, in terms of its placement of the object masked in red?
Describe the scene and describe the object masked in red (what is it? Where should the object be according to the objective?), then respond with 'first' or 'second' in json:

```json
{
  "final_answer": "A"/"B"
}
```
\end{lstlisting}

The results generated with Gemini 1.5Pro for text inputs is shown in \Cref{tab:text_vs_baseline} and the result for image inputs (See \Cref{sec:image_inputs}) is shown in \Cref{tab:image_vs_baseline}, showing \methodname's superior performance over both baselines in both task settings.

\begin{table}[tb]
    \centering
    \begingroup
    \setlength{\tabcolsep}{5pt} 
    \renewcommand{\arraystretch}{1.} 
    \begin{tabular}{@{\hskip 0.5em}rcccc@{\hskip 0.5em}}
    \toprule
    & vs. LayoutGPT & vs. Holodeck & \\
    \midrule 
    LayoutGPT wins & - & 0.56  \\
    Holodeck wins & 0.44 & -  \\
    Ours wins & \textbf{0.72} & \textbf{0.72} \\
    \hline
    \end{tabular}
    \caption{Win-rate comparison between our method and LayoutGPT and Holodeck according to Gemini 1.5Pro as judge, for placement tasks with image input.}
    \label{tab:image_vs_baseline}
    \endgroup
\end{table}

\begin{table}[tb]
    \centering
    \begingroup
    \setlength{\tabcolsep}{5pt} 
    \renewcommand{\arraystretch}{1.} 
    \begin{tabular}{@{\hskip 0.5em}rcccc@{\hskip 0.5em}}
    \toprule
    & vs. LayoutGPT & vs. Holodeck & \\
    \midrule 
    LayoutGPT wins & - & 0.54  \\
    Holodeck wins & 0.46 & -  \\
    Ours wins & \textbf{0.70} & \textbf{0.72} \\
    \hline
    \end{tabular}
    \caption{Win-rate comparison between our method and LayoutGPT and Holodeck according to Gemini 1.5Pro as judge, for placement tasks with text input.}
    \label{tab:text_vs_baseline}
    \endgroup
\end{table}


\begin{figure*}
    \centering
    \noindent\begin{mdframed}
    \begin{lstlisting}[escapechar=@]
The desired placement of the object masked in red can be described by:
@\colorbox{light-gray}{[Placement text prompt]}@

The predicted placement of the object is shown in the predicted image below:
@\colorbox{light-gray}{[Masked rendering of predicted final placement,}@
@\colorbox{light-gray}{with placed object in semi-transparent red mask.]}@

Please grade the predicted placement of the predicted placement of the object masked in red on a scale of 1-4, where:
1 = either the target object is not observed at all within the scene, or the placement is physically implausible (e.g. the object is floating in the air or intersecting with other objects)
2 = The object is observed within the scene, but its placement differs substantially from the placement observed in the reference image.
3 = The object is observed within the scene, and its placement is sensible (e.g. the target object is NOT floating in air or intersecting with other objects), and the placement of the target object is  different from the reference image, but not in substantial ways.
4 = a totally sensible placement of the target object, and captures both valid physics (e.g. the target object is NOT floating in air or intersecting with other objects) as well as considerations for function, accessibility, and aesthetics.

Please reason about what is in 
Describe the scene and describe the object masked in red (what is it? Where should the object be according to the objective?), then respond with the final score in json:

```json
{
  "final_answer": 1/2/3/4
}
```
    \end{lstlisting}
    \end{mdframed}
    \caption{Prompt for extracting plausibility scores.}
    \label{fig:plausibility_prompt}
\end{figure*}

\begin{figure*}
    \centering
    \noindent\begin{mdframed}
    \begin{lstlisting}[escapechar=@]
  * Contact (surface1, surface2)
    Enforces that surface1 and surface2 are in contact with each other.
  * FarFrom (surface1, surface2, const)
    This enforcs that surface1 and surface2 are at a AT LEAST some distance away from eachother specified by const, a float in CENTIMETERS in the life-size 3D scene.
  * CloseTo (surface1, surface2, const)
    This enforcs that surface1 and surface2 are AT MOST some distance away from eachother specified by const, a float in CENTIMETERS in the life-size 3D scene.
  * Parallel (surface1, surface2)
    This enforces that surface1 and surface2 are parallel.
  * Above (surface1, surface2)
    This enforces that surface1 is ABOVE surface2 (note the order.) This does NOT ensure that the birdseye view of object1 and object2 overlap.
  * NoOverhang (surface1, surface2)
    This enforces that surface1's vertical projection is entirely contained in surface2's projection. This is good when you'd like one surface to be entirely contained within another, for instance for physical stability.
    Also, if you ever want to ensure contact with floor, make sure to use NoOverhang as well.
    \end{lstlisting}
    \end{mdframed}
    \caption{Prompt for constraint documentation}
    \label{fig:constraint_documentation}
\end{figure*}

\begin{figure*}
    \centering
    \noindent\begin{mdframed}
    \begin{lstlisting}[escapechar=@]
@\colorbox{light-gray}{[Source layout rendering]}@

The following text describes the target object that we would like to place in the image above: @\colorbox{light-gray}{[Placement text prompt]}@
In each of the reference images below, a target object is masked in red. Even though this may be different from the target object just mentioned and even though the scene may be different, the placements of the object shares underlying patterns and structure across the scenes. 

You are given a library of the following functions:
above: @\colorbox{light-gray}{[Constraint library doc string]}@

For the {item_to_focus_on}, return a json that has a list. Each element of the list specifies:
1) in the field called 'constraints': the constraint function (e.g. Parallel)
2) in the field called 'surface1' a string describing surface1 -- if it's a surface on the target object (or collection of objects), refer to it as the 'target object'. (e.g. bottom of the target object)
3) in the field called 'surface2' a string describing surface2 -- if it's a surface on the target object (or collection of objects), refer to it as the 'target object'. (e.g. top of the table in the corner of the room)

Return a json marked by ```json at the very beginning, with a list of ALL of the constraints that is relevant to the {item_to_focus_on}.
NOTE: since none of the objects are floating in the scene, ALWAYS start off the list with a CONTACT constraint. What is the target object in contact with? Which surface of the target object is in contact with something else?

NOTE: specify distance relations between objects using the constraints `CloseTo` and `FarFrom`, by spcifying which surfaces on the target object and the surounding objects are relevant to these distance relations.
For instance, if the target object is a potted plant at the corner of the room, you would use the CloseTo distance relation with respect to one side (left or right) of the plant, the surface of the correct wall, and another CloseTo distance relation between the back side of the plant and the other wall forming the corner.
Alternatively, if the target object is something mounted on the wall above something else, it would make sense to specify FarFrom constraint between the bottom surface of the target object and the surface of the object underneath.
Note that in both cases, you must specify a threshold distance in Centimeters. Use the assumption that the scene is real-life sized in 3D.

You are allowed to return an empty list if nothing is relevant to the description of the desired physical relation above.

    \end{lstlisting}
    \end{mdframed}
    \caption{Prompt for constraint outline generation}
    \label{fig:constraint_outline_generation}
    \vspace{2em}
\end{figure*}

\begin{figure*}
    \centering
    \noindent\begin{mdframed}
    \begin{lstlisting}[escapechar=@]
The following text describes the target object to be placed into the scene: 
@\colorbox{light-gray}{[Placement text prompt]}@

The target object is related to another 'anchor' object within the scene by the physical relation: @\colorbox{light-gray}{[Constraint outline]}@
The surface of that 'anchor' object in the physical relation can be described by: @\colorbox{light-gray}{[constraint outline]}@
We refer to this object as the anchor object.
    
In the following image, I want you to find the anchor object among the segmentation masks I show in different colors.
What is the color of segmentation mask of the the anchor object in the following image?[@\colorbox{light-gray}{concat color names}@/none]

@\colorbox{light-gray}{[Insert masked rendering of anchor objects]}@

I want you to reason about it, then output a json, like:
```json
{'final_answer': "@\colorbox{light-gray}{concat color names}@/none"
    }
```
respond with `none` if none of the segmentation masks in the image match the anchor masked object.
    \end{lstlisting}
    \end{mdframed}
    \caption{Prompt for extracting anchor objects}
    \label{fig:extract_anchor_object}
\end{figure*}

\begin{figure*}
    \centering
    \noindent\begin{mdframed}
    \begin{lstlisting}[escapechar=@]
The following image shows our "anchor" object in a semi-transparent red mask.

@\colorbox{light-gray}{[insert masked rendering of anchor object]}@

Here's a 3D plot of the @\colorbox{light-gray}{[ANCHOR/TARGET]}@ object in its canonical space. 
In this plot:
The direction 'upwards' is described by the positive z direction. (down is negative)
the direction 'right' is described by the positive x direction. (left is negative)
the direction 'forwards' id described by the negative y direction. (backwards is positive)

@\colorbox{light-gray}{[Insert 3D plot of ANCHOR/TARGET object ]}@

And the following text describes the target object: @\colorbox{light-gray}{[Placement text prompt]}@

You've decided the following must be true:
@\colorbox{light-gray}{[Constraint outline]}@

Think about the @\colorbox{light-gray}{[ANCHOR/TARGET]}@ object. Which object does that correspond to in the above?
Think about the surface where the interaction between the two objects is happening. Which way is it pointing on the @\colorbox{light-gray}{[ANCHOR/TARGET]}@ object?

I'd like you now to tell me how you would extract the relevant surfaces from our @\colorbox{light-gray}{[ANCHOR/TARGET]}@ object relevant for the placements described above. The definition of a relevant surface is one that is involved in physical constraints. If something is supposed to be to the RIGHT of this object, then it makes sense that a surface pointing to the right is  relevant, since that surface can be usd to judge whether the object is truly to the right of the object. 
If the anchor object were a table, and the target object is to be put onto the table then the relevant surface of interaction for the ANCHOR OBJECT is pointing UPWARDS. The surface of interaction for the TARGET OBJECT is pointing downwards.
The surface you extract does NOT NEED TO BE in physical contact with the other object. For instance, if a distance is to be maintained between two objects, think which surface is most relevant in that distance calculation.

I want you to reason about it, then output a json to specify the direction of the surface in the @\colorbox{light-gray}{[ANCHOR/TARGET]}@ object relevant to the interaction, like:


```json
{'final_answer': up/down/left/right/front/back}
```
You cannot return the word 'none'.

    \end{lstlisting}
    \end{mdframed}
    \caption{Prompt for extracting object surface directions.}
    \label{fig:extract_surfaces}
\end{figure*}

\begin{figure*}
    \centering
    \noindent\begin{mdframed}
    \begin{lstlisting}[escapechar=@]
The following image shows our "anchor" object in a semi-transparent red mask.

@\colorbox{light-gray}{[Insert masked rendering of ANCHOR object]}@

Here's a 3D plot of the @\colorbox{light-gray}{[ANCHOR/TARGET]}@ object in its canonical space. 
In this plot:
The direction 'upwards' is described by the positive z direction. (down is negative)
the direction 'right' is described by the positive x direction. (left is negative)
the direction 'forwards' id described by the negative y direction. (backwards is positive)

@\colorbox{light-gray}{[Insert 3D plot of ANCHOR/TARGET object]}@

And the following text describes the target object: @\colorbox{light-gray}{[Placement text prompt]}@

You've decided the following must be true:
@\colorbox{light-gray}{[Constraint outline]}@

Think about the @\colorbox{light-gray}{[ANCHOR/TARGET]}@ object. Which object does that correspond to in the above?
Think about the surface where the interaction between the two objects is happening. Which surface (@\colorbox{light-gray}{[LIST OF COLORS]}@) is relevant to the interaction?

I want you to reason about it, then output a json to specify the the surface in the @\colorbox{light-gray}{[ANCHOR/TARGET]}@ object relevant to the interaction, like:

```json
{'final_answer': "@\colorbox{light-gray}{[LIST OF COLORS]}@/none"
    }
```
    \end{lstlisting}
    \end{mdframed}
    \caption{Prompt for choosing object surfaces based on the color of the surface in the visualization.}
    \label{fig:choose_surface}
\end{figure*}

\begin{figure*}
    \centering
    \noindent\begin{mdframed}
    \begin{lstlisting}[escapechar=@]
    
The following image shows our "anchor" object in a semi-transparent red mask.

@\colorbox{light-gray}{[Insert masked rendering of ANCHOR object]}@

Here's a 3D plot of the @\colorbox{light-gray}{[ANCHOR/TARGET]}@ object in its canonical space. 
In this plot:
The direction 'upwards' is described by the positive z direction. (down is negative)
the direction 'right' is described by the positive x direction. (left is negative)
the direction 'forwards' id described by the negative y direction. (backwards is positive)

@\colorbox{light-gray}{[Insert 3D plot of ANCHOR/TARGET object]}@

And the following text describes the target object: @\colorbox{light-gray}{[Placement text prompt]}@

You've decided the following must be true:
@\colorbox{light-gray}{[constraint outline]}@

In order to enforce this constraint, there's a few parameters you must specify.
Use your visual judgment to determine the best values for the following parameters:
@\colorbox{light-gray}{[Arg documentation string for each constraint function]}@

For the target object, you chose the surface visualized below:

@\colorbox{light-gray}{[Colorized visualization of the target object and the interaction surface chosen]}@

For the anchor object, you chose the surface visualized below:

@\colorbox{light-gray}{[Colorized visualization of the anchor object and the interaction surface chosen]}@

What approximation of the parameters make the most sense, given the surfaces that you've chosen, and the constraint you've chosen to enforce?
Return a json with each value specified in the json. Begin the json with ```json.

    \end{lstlisting}
    \end{mdframed}
    \caption{Prompt for estimating continuous parameters}
    \label{fig:estimate_parameters}
\end{figure*}

\begin{figure*}
    \centering
    \noindent\begin{mdframed}
    \begin{lstlisting}[escapechar=@]
You are an experienced room designer.
  
Please help me arrange objects in the room by assigning constraints to each object.
Here are the constraints and their definitions:

1. distance constraint:
    1) near, object: near to the other object, but with some distance, 50cm < distance < 150cm.
    2) far, object: far away from the other object, distance >= 150cm.
2. position constraint:
    1) in front of, object: in front of another object.
    2) around, object: around another object, usually used for chairs.
    3) side of, object: on the side (left or right) of another object.
    4) left of, object: to the left of another object.
    5) right of, object: to the right of another object.
    6) behind of, object: behind another object.
    7) in front of, object: in front of another object.
    8) ontop of, object: on top of another object. 
3. direction constraint:
    1) face to, object: facing another object.
4. alignment constraint:
    1) center aligned top, object: align the center of the object with the center of the TOP of another object.
    2) center aligned front, object: align the center of the object with the center of the FRONT of another object.
    3) center aligned side, object: align the center of the object with the center of the SIDE of another object.
    \end{lstlisting}
    \end{mdframed}
    \caption{Prompt for Holodeck (Part 1) -- For part 2, see \Cref{fig:baseline_holodeck_prompt2}.}
    \label{fig:baseline_holodeck_prompt1}
\end{figure*}

\begin{figure*}
    \centering
    \noindent\begin{mdframed}
    \begin{lstlisting}[escapechar=@]
For each object, you can select various numbers of constraints and any combinations of them and the output format must be:
object | constraint 1 | constraint 2 | ...
For example:
coffee table-0 | near, sofa-0 | in front of, sofa-0 | center aligned front, sofa-0 | face to, sofa-0
tv stand-0 | far, coffee table-0 | in front of, coffee table-0 | center aligned front, coffee table-0 | face to, coffee table-0
desk-0 | far, tv stand-0
chair-0 | in front of, desk-0 | near, desk-0 | center aligned front, desk-0 | face to, desk-0
floot lamp-0 | near, chair-0 | side of, chair-0

Here are some guidelines for you:
1. The objects of the *same type* are usually *aligned*.
2. When handling chairs, you should use the around position constraint. Chairs must be placed near to the table/desk and face to the table/desk.

In the above examples, "coffe table-0", "sofa-0", "tv stand-0", "desk-0", "chair-0", "floot lamp-0" are all object IDS 
in the scene. In reality, the object IDs look more like "object-0", "object-1", ...
i.e. object-1 | in front of, object-2 | near, object-2 | face to, object-3

Here is a list of preexisting objects in the scene, in the format  [object_id]: [object_description]
@\colorbox{light-gray}{[Descriptions and labels of preexisting objects in the scene]}@ 

Here is the object that I want to place in the room (object id of it is object-target):
@\colorbox{light-gray}{[Description of transformable object]}@ 
Please first use natural language to explain your high-level design strategy, and then follow the desired format *strictly* (do not add any additional text at the beginning or end) to provide the constraints for the object.
    \end{lstlisting}
    \end{mdframed}
    \caption{Prompt for Holodeck (Part 2) -- For Part 1, see \Cref{fig:baseline_holodeck_prompt1}.}
    \label{fig:baseline_holodeck_prompt2}
\end{figure*}

\begin{figure*}
    \centering
    \noindent\begin{mdframed}
    \begin{lstlisting}[escapechar=@]
Instruction: synthesize the 3D layout of an indoor scene. The generated 3D layout should follow the CSS style, where each line starts with the furniture category and is followed by the 3D size, orientation and absolute position. Formally, each line should follow the template:
FURNITURE {length: ?cm; width: ?cm; height: ?cm; left: ?cm; top: ?cm; depth: ?cm; orientation: ? degrees;}
All  values are in cm but the orientation angle is in degrees.

Here are the info of other objects within the scene:
@\colorbox{light-gray}{[List of objects inside the scene, colon separated from}@
@\colorbox{light-gray}{their width, height, left, top, depth and orientation information.]}@ 

Generate a line for an object described by the following:
@\colorbox{light-gray}{[Description of transformable object]}@ 
    \end{lstlisting}
    \end{mdframed}
    \caption{Prompt for LayoutGPT}
    \label{fig:baseline_layoutgpt_prompt}
\end{figure*}

\end{document}